\documentclass[11pt]{article}
\usepackage{natbib}
\usepackage{amsmath,amstext,amssymb,amsfonts,amscd,color}
\usepackage{graphicx}
\usepackage{enumitem,epsfig}
\usepackage[hmargin={1.0in, 1.0in}, vmargin={1.2in, 1.2in}]{geometry}
\usepackage{threeparttable}
\usepackage{cases,color}
\usepackage[table,dvipsnames]{xcolor}
\usepackage{verbatim}
\usepackage{booktabs}
\usepackage{multirow}
\usepackage{multicol}
\usepackage{adjustbox}
\usepackage{diagbox}
\usepackage{bbm}
\usepackage{amsthm}
\usepackage{bm}
\usepackage{authblk}
\usepackage{subcaption}
\usepackage{algorithm}
\usepackage{algpseudocode}

\usepackage[
  colorlinks=true,
  citecolor=blue,
  urlcolor=blue,
  linkcolor=blue,
  filecolor=blue]{hyperref}
\usepackage{cleveref}
\allowdisplaybreaks[4]

\usepackage{xr-hyper}
\makeatletter
\newcommand*{\addFileDependency}[1]{
  \typeout{(#1)}
  \@addtofilelist{#1}
  \IfFileExists{#1}{}{\typeout{No file #1.}}
}
\makeatother
\newcommand*{\myexternaldocument}[1]{%
    \externaldocument{#1}%
    \addFileDependency{#1.tex}%
    \addFileDependency{#1.aux}%
}
\myexternaldocument{arxiv_supp_JASA_Initial_placeholder}

\newtheorem{thm}{Theorem}[section]
\newtheorem{cor}{Corollary}[section]

\newtheorem{rmk}{Remark}[section]

\newtheorem{defi}{Definition}[section]

\newtheorem{prop}{Proposition}[section]
\newtheorem{assumpt}{Assumption}[section]

\numberwithin{equation}{section}
\numberwithin{figure}{section}
\numberwithin{table}{section}

\newcommand{\cb}{\mathcal{B}}
\newcommand{\cc}{\mathcal{C}}
\newcommand{\cd}{\mathcal{D}}

\newcommand{\ci}{\mathcal{I}}

\newcommand{\cn}{\mathcal{N}}

\newcommand{\cR}{\mathcal{R}}

\newcommand{\ct}{\mathcal{T}}

\newcommand{\br}{\mathbb{R}}
\newcommand{\bz}{\mathbb{Z}}

\newcommand{\fone}{\mathbf{1}}

\newcommand{\Multi}{\mbox{Multinomial}}

\newcommand{\blrp}[1]{\mathbb{P}\left({#1}\right)}

\newcommand{\lrp}[1]{\left({#1}\right)}

\newcommand{\lrcp}[1]{\left\{{#1}\right\}}

\newcommand{\bigag}[1]{\big\langle{#1}\big\rangle}

\newcommand{\BE}{\begin{equation}}
\newcommand{\EE}{\end{equation}}
\newcommand{\BEqn}{\begin{eqnarray*}}
\newcommand{\EEqn}{\end{eqnarray*}}
\newcommand{\benu}{\begin{enumerate}[label=(\roman*)]}
\newcommand{\eenu}{\end{enumerate}}

\DeclareMathOperator*{\argmax}{argmax}
\DeclareMathOperator*{\argmin}{argmin}

\def\spacingset#1{\renewcommand{\baselinestretch}%
{#1}\small\normalsize}
\spacingset{1.25}

\begin{document}

\title{Dynamic Topic Modeling with a Higher-Order Hypergraphical Representation}
\author[1]{Hanjia Gao}
\author[1]{Hanwen Ye}
\author[2]{Qing Nie}
\author[3]{Annie Qu}

\affil[1]{Department of Statistics, University of California, Irvine}
\affil[2]{Department of Mathematics and Department of Developmental \& Cell Biology, University of California, Irvine}
\affil[3]{Department of Statistics and Applied Probability, University of California, Santa Barbara}
\date{}	
\maketitle

\begin{abstract}

Dynamic topic modeling is widely used to analyze evolving trends in scientific literature, medical records, and social media. Traditional topic models represent each topic through a single probability vector on the multinomial simplex and implicitly couple word occurrence and repetition within one probabilistic mechanism. However, this formulation restricts the dependence structure among words and overlooks informative higher-order interactions, particularly in dynamic corpora with overlapping semantics.
To address these limitations, we introduce a hypergraphical representation of text where each document is modeled as a hyperedge connecting all co-occurring words, with repetition intensities encoded as node weights. This representation naturally separates word occurrence from repetition and induces a novel hypergraph-based multinomial distribution with a nonlinear normalization depending on the observed word set of each document. Building on this likelihood, we develop a dynamic topic modeling framework via structured low-rank factorizations with explicit temporal regularization on topic–word profiles. 
Moreover, in theory, we establish local convergence guarantees and derive non-asymptotic error bounds despite the intrinsic nonconvexity of bilinear factorization and document-specific nonlinear normalization.
Numerical experiments on synthetic data and an application to the International Conference on Learning Representations (ICLR) corpus demonstrate consistent improvements over existing multinomial-based topic models.
\end{abstract}

\noindent%
\vfill

\newpage
\spacingset{1.8} 

\section{Introduction}

Topic modeling aims to uncover latent semantic structure and quantify topic prevalence in large document corpora. It has become a central tool in statistical text analysis, with applications in scientific literature tracking, medical record analysis \citep{sarioglu2012clinical, ye2024dynamic}, social media monitoring \citep{curiskis2020evaluation}, and e-commerce archives \citep{palese2018relative, yuan2018topic}. In many contemporary settings, corpora are collected over extended time horizons, and both topic-word profiles and document–topic associations evolve dynamically. This temporal heterogeneity motivates the development of dynamic topic models capable of tracking structural changes in latent topics over time.

Classical topic modeling methods apply multinomial likelihood to the bag-of-words (BOW) representation \citep{harris1954distributional} of documents, encoding documents by marginal word counts. Two dominant frameworks within this probabilistic paradigm are Latent Dirichlet Allocation (LDA) \citep{blei2003latent} and probabilistic latent semantic indexing (pLSI) \citep{hofmann1999probabilistic}. LDA-based methods \citep{blei2006correlated, roberts2014structural, chen2020modeling, sobhani2024multi, mcauliffe2007supervised, chong2009simultaneous} adopt Bayesian formulations and are typically fitted via variational inference with state-space chaining and time-invariant prior in dynamic extensions \citep{blei2006dynamic}. Meanwhile, pLSI-based approaches \citep{arora2012learning, arora2013practical, klopp2023assigning, ke2024using} mainly focus on static settings, formulating topic modeling as low-rank matrix factorization and exploiting geometric or spectral properties of word distributions under separability assumptions.

However, despite their wide applications, both frameworks apply the multinomial likelihood to BOW counts. Conditioning on document length, word tokens are drawn independently from a single simplex-constrained probability vector, and all co-occurrence structure is determined by marginal composition. This modeling choice has three important implications. First, the dependence structure within each document is fully determined by the topic-specific probability vector and cannot adapt to document-specific word interaction patterns. Second, word occurrence and repetition may exhibit distinct behaviors but are coupled through the same parameter. Third, topic identifiability relies solely on marginal word distributions, weakening separability when topics overlap in marginals but differ in joint occurrence–repetition patterns.

To capture document-specific higher-order dependencies and decouple word occurrence from repetition, we consider a novel hypergraph representation of documents. Specifically, we treat each vocabulary term as a node and each document as a weighted hyperedge supported on the words appearing in the document. The support of each hyperedge captures the document's word co-occurrence pattern, while the node weights encode word repetition intensities. For example, in a corpus of data science articles, dataset-centric papers often focus on a single benchmark dataset, forming a hyperedge heavily weighted on one dataset node (i.e., terms like ``ImageNet32'' and ``MNIST''), whereas methodology papers reference multiple datasets for data illustration, leading to hyperedges connected to many nodes but with lower weights per node. Thus, unlike BOW, this representation decouples word occurrence from repetition through hyperedge support and node weights, allowing interaction patterns to vary across document-specific word subsets rather than being governed by a single multinomial parameter. By leveraging joint activation and repetition patterns beyond marginal proportions, it provides additional discriminatory information and improves topic identifiability when topics overlap in semantic content.

Building on this representation, we develop a dynamic topic modeling framework that models word occurrence and repetition separately while allowing topic structure to evolve over time. Specifically, we model the support of each hyperedge through a Bernoulli component and, conditioning on this support, model node weights via a multinomial distribution with document-specific normalization. Leveraging a mixed-membership formulation, we impose structured low-rank factorizations on both components to characterize document–topic associations and topic–word profiles.
Under standard nonnegativity and simplex constraints, the low-rank factor can be naturally interpreted as topic–document associations and topic–word behaviors (i.e., occurrence and repetition), offering direct insight into how topic prevalence and semantic usage evolve over time.
In dynamic corpora, topics are expected to evolve gradually while maintaining coherent themes. For example, as research on ``language models'' develops, word usage patterns of a paper may shift, yet the overall thematic focus remains on ``language models.'' To accommodate this setting, we impose temporal regularization directly on topic–word profiles, allowing smooth semantic drift while preserving identifiability.

Our contributions are threefold. 
First, we introduce a hypergraph-based probabilistic representation for text corpora that captures higher-order dependence structures beyond the multinomial model and explicitly separates word occurrence from repetition intensity. 
Second, we develop a dynamic modeling framework that accommodates temporally evolving corpora through structured low-rank factorization with direct temporal regularization, providing a likelihood-based alternative for dynamic topic modeling.
Third, despite the intrinsic nonconvexity induced by bilinear factorization and document-specific normalization, we establish local convergence guarantees and derive explicit non-asymptotic Frobenius-norm error bounds by developing novel perturbation and concentration arguments. 

The remainder of the article is organized as follows. Section \ref{Sec:Representation} introduces the hypergraph representation and the induced distribution. Section \ref{Sec:Method} presents the dynamic modeling framework and estimation algorithm. Theoretical properties are established in Section \ref{Sec:Theory}. Numerical experiments are reported in Section \ref{Sec:NumericalStudies}. Section \ref{Sec:Conclusion} concludes with a discussion and directions for future work.

\section{Text Representation via Hypergraph}\label{Sec:Representation}

Under BOW representation, a document is encoded as a count vector $d=(d_1,\cdots,d_p)$, where $p$ is the number of unique words (i.e., vocabulary size) and $d_j$ denotes the total number of occurrences of word $j$. Most likelihood-based topic models assume that $d$ follows a multinomial distribution with a $p$-dimensional compositional parameter that quantifies the event probabilities. Consequently, each word token is treated as conditionally independent, and the entire joint distribution of word counts is determined by a single compositional vector, limiting document-level word interactions to those implied by marginal proportions.

To encode document-specific interaction structure beyond marginal composition, we represent each document as a weighted hyperedge in a hypergraph. While graphs have been used to model pairwise word co-occurrences \citep{rousseau2013graph, rousseau2015text, yao2019graph}, they are restricted to pairwise interactions since each edge of a graph connects only two nodes. 
A hypergraph generalizes a graph by allowing each hyperedge to connect an arbitrary subset of nodes, and therefore, directly encodes document-level co-occurrence structure. Recent works incorporate hypergraphs into neural architectures for text analysis \citep{ding2020more, pradeepa2024hgatt_lr, bazaga2024hyperbert}, primarily as architectural enhancements. In contrast, we employ hypergraphs as an explicit probabilistic representation of word interactions.

\begin{figure}
    \centering
    \includegraphics[width=.95\linewidth]{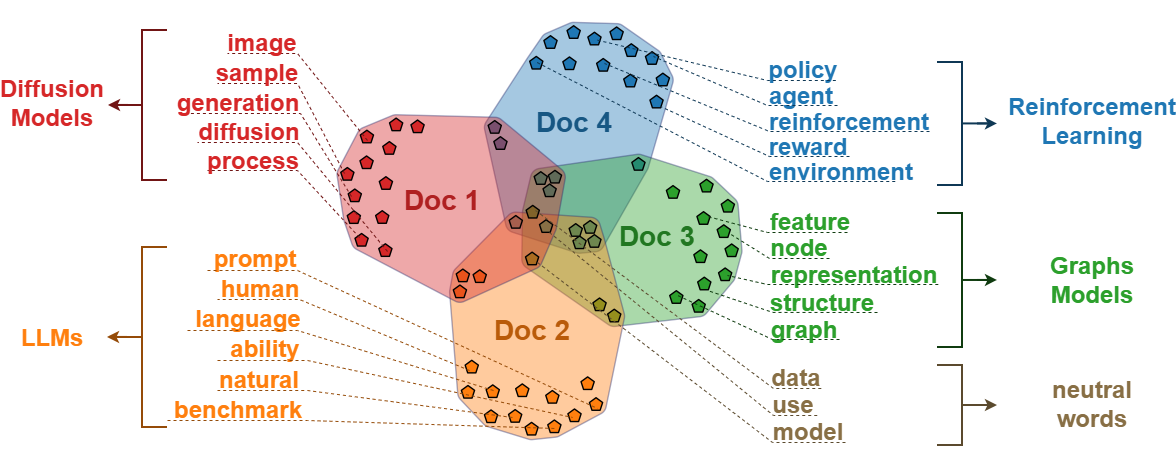}
    \caption{Illustration of the hypergraphical representation of text documents. Each marker represents a node (word), and each colored region represents a hyperedge (document).}
    \label{Fig:Illustration}
\end{figure}

Given a corpus with vocabulary size $p$, we construct a hypergraph with $p$ nodes, each corresponding to a vocabulary word. Each document is represented by a hyperedge supported on the set of words appearing in it, and node weights record repetition intensities within that document. The hyperedge support captures word activation and co-occurrence patterns, while node weights quantify heterogeneous repetition behavior. An illustrative example is provided in Figure \ref{Fig:Illustration}, where each colored region represents a hyperedge. For visual clarity, repetition intensities are not displayed. We observe that hyperedge-specific words reveal the semantic theme of each hyperedge, whereas overlapping words reflect shared activation patterns across documents.

Formally, each document is represented by a pair $(e,r)$. The binary vector $e=(e_1,\cdots,e_p)^\top \in \{0,1\}^p$ encodes hyperedge support, where $e_j=1$ if word $j$ appears in the document and $e_j=0$ otherwise. The repetition vector $r=(r_1,\cdots,r_p) \in \bz_{\geq0}^p$ records additional occurrences after the first appearance, so that the observed count vector satisfies $d=e+r$. This decomposition separates word occurrence from repetition intensity, linking them only through document-specific support.
The construction of $e$ follows the convention in \citep{wu2024general}, while the construction of $r$ is novel, as heterogeneous node weights are typically ignored in existing hypergraph analyses.

We model word activation and repetition through a two-layer mechanism. Motivated by conditional independence assumptions in network models \citep{holland1983stochastic, hoff2002latent, ke2019community, wu2024general}, we assume that $e_j$ are independently generated from Bernoulli distributions with parameters $q_j \in (0,1)$ across words. Conditional on $e$, repetition counts $r_j$ are independently generated from a Poisson distribution with parameter $\lambda_j > 0$ if $e_j=1$, and $r_j=0$ otherwise. The Poisson layer provides an exponential-family construction that yields independent repetition counts across activated words before conditioning on the total repetition count.

Let $s = \sum_j r_j$ denote the total repetition count. Conditional on $(e,s)$, the repetition vector $r$ follows a multinomial distribution, that is
\begin{equation*}
    r ~|~ (e, s) \sim \Multi(s, \theta(e)), \qquad
    \theta_j(e) = \frac{\lambda_j e_j}{\sum_u \lambda_u e_u}
\end{equation*}
Thus, repetition intensities are normalized over the activated support $\{j:e_j=1\}$. Since $\lambda$ influences $\theta$ only through relative ratios, we impose the normalization $\sum_j \lambda_j = p$ for identifiability. Recall that $d=e+r$, this construction induces a distribution of $D$ as formalized below.

\begin{defi}[Hypergraph-induced multinomial distribution]\label{Def:H_Multinomial}
Let $s\in\mathbb{Z}_{\ge0}$, $q\in(0,1)^p$, and $\lambda\in\mathbb{R}_+^p$ with $\sum_j \lambda_j=p$. For any vector $d\in\bz_{\geq0}^p$, define $e_j(d)=\fone\{d_j>0\}$, $r_j(d)=d_j-e_j(d)$ and $\tilde\cd = \{d\in\bz_{\geq0}^p: ~\sum\nolimits_j e_j(d) > 0, ~\sum\nolimits_j r_j(d) = s\}$.
We say $D$ follows a hypergraph-induced multinomial (H-Multinomial) distribution with parameters $(s,q,\lambda)$ if
\begin{equation*}
    \blrp{D=d}
  = \left\{
    \begin{array}{ll}
        \displaystyle\binom{s}{r_1 \cdots r_p}
        \lrp{\prod\limits_{j=1}^{p} {q_j}^{e_j(d)} (1-q_j)^{1-e_j(d)}} 
        \lrp{\prod\limits_{j=1}^{p} 
             \lrp{\frac{\lambda_j e_j(d)}
                       {\sum_t \lambda_t e_t(d)}}^{r_j(d)}}, 
         & d\in\tilde\cd, \\
         \prod\nolimits_{j=1}^{p} (1-q_j),
         & d = \mathbf{0}, \\
         0, & \mbox{otherwise}.
    \end{array}
    \right.
\end{equation*}
\end{defi}

Throughout, the H-Multinomial distribution refers to the conditional distribution of $D$ given $s$, with $s$ modeled separately in Section \ref{Sec:Method}. Marginally, the activation vector $e$ follows a product Bernoulli distribution, and conditioning on $e$ and $s$, repetition counts are multinomially allocated over the activated support. The induced distribution is therefore a mixture of multinomials indexed by support patterns, with mixing weights determined by the Bernoulli activation layer. Note that when all words are activated (i.e., $e_j=1$ for all $j$), the H-Multinomial reduces to a standard multinomial distribution.
Moreover, Proposition \ref{Prop:validity} justifies that the H-Multinomial distribution is well-defined and is uniquely characterized by the parameters $(s,q,\lambda)$.

\begin{prop}\label{Prop:validity}
For any $s \in \bz_{\geq0}$, $q \in (0,1)^p$, and $\lambda \in \mathbb{R}_+^p$ with $\sum_j \lambda_j = p$, the H-Multinomial distribution is a valid probability distribution on $\mathbb{Z}_{\ge0}^p$. Moreover, if $s\ge1$ and two parameter sets $(s,q,\lambda)$ and $(\tilde{s},\tilde{q},\tilde{\lambda})$ induce the same distribution, then $s=\tilde{s}$, $q=\tilde{q}$, and $\lambda=\tilde{\lambda}$.
\end{prop}

Unlike the standard multinomial distribution, where word occurrence and repetition are jointly governed by a single compositional vector that is fixed across documents, the H-Multinomial parametrically decouples occurrence probabilities $q$ from repetition parameters $\lambda$, while statistically coupling repetition counts through support-dependent normalization. Through such a support-dependent normalization in the multinomial layer, activation patterns in $e$ directly influence repetition allocation, thus dependence among repetition counts varies across activated subsets rather than being governed by a universal parameter. This support-dependent structure provides additional discriminatory information when topics exhibit overlapping marginal proportions.

\section{Dynamic Topic Modeling}\label{Sec:Method}

In this section, we develop a dynamic topic modeling framework based on the proposed H-Multinomial distribution. 
Section \ref{Sec:Method_Motivation} introduces the modeling goals and design principles. 
Building on these principles, Section \ref{Sec:Method_Model} formalizes the low-rank structure of the dynamic model and identifiability conditions.
Section \ref{Sec:Method_Estimation} then formulates the penalized likelihood problem, and Section \ref{Sec:Method_Algorithm} presents an efficient estimation algorithm.

\subsection{General Principles}\label{Sec:Method_Motivation}

Our dynamic topic model is guided by three principles.

First, the H-Multinomial distribution separates word occurrence probabilities $q$ and repetition intensities $\lambda$. These components encode distinct linguistic signals that are conflated under the single compositional parameter in the BOW-multinomial representation, and we model them through separate mechanisms.

Second, both occurrence probabilities and repetition intensities share low-rank topic structure. As in classical mixed-membership models \citep{hofmann1999probabilistic, blei2003latent}, each document is represented as a mixture of latent topics. The document-level topic weights jointly determine both occurrence and repetition behavior. The two signals provide complementary information for distinguishing topics while remaining semantically tied through the shared mixture.

Third, topic semantics are allowed to evolve smoothly over time after temporal alignment. While document–topic proportions can vary freely across time windows, topic–word profiles are assumed to fluctuate moderately around their temporal averages.

\subsection{Model Specification via Low-Rank Factorization}\label{Sec:Method_Model}

Let $\cd = \{\cd_t\}_{t=1}^{T}$ denote the corpora observed over $T$ time windows, where $\cd_t = \{d_{ti}\}_{i=1}^{n_t}$ with a vocabulary size of $p$. For each document, define the binary activation vector $e_{ti} = (\fone\{d_{ti1}>0\}, \cdots, \fone\{d_{tip}>0\})^{\top}$, the repetition vector $r_{ti} = d_{ti} - e_{ti}$, and total repetition count $s_{ti} = \sum_j r_{tij}$. We assume that $d_{ti}$'s are all independent, and each $d_{ti}$ follows the H-Multinomial distribution with true parameters $(s_{ti}, q_{ti}^{\ast}, \lambda_{ti}^{\ast})$. Let
\begin{equation*}
    Q_t^{\ast} = [q_{t1}^{\ast}; \cdots; q_{tn_t}^{\ast}]^{\top} \in (0,1)^{n_t \times p}, \qquad
    \Lambda_t^{\ast} = [\lambda_{t1}^{\ast}; \cdots; \lambda_{tn_t}^{\ast}]^{\top} \in \br_{+}^{n_t\times p}
\end{equation*}
collect the corresponding parameters across documents. 
Moreover, we model the total repetition counts of each document through the activated support of words.

\begin{assumpt}[Repetition scaling]\label{Assumpt:rho}
For each document $(t,i)$, we assume
\begin{equation*}
    \sum\nolimits_{j=1}^{p} e_{tij} \geq 1,
    \qquad
    s_{ti} = \sum\nolimits_{j=1}^{p} r_{tij} 
    = \rho_{ti} \sum\nolimits_{j=1}^{p} e_{tij},
\end{equation*}
where $\rho_{ti} \in [\underline\rho, \bar\rho]$ for fixed constants $0 < \underline{\rho} \le \bar{\rho} < \infty$.
\end{assumpt}

This assumption links total repetition intensity to document vocabulary richness 
and ensures document lengths remain uniformly controlled, which is consistent with empirical observations.

Suppose there are $K$ distinct topics overall, and let $K_t$ denote the number of active topics present at time $t$. We assume that $(Q_t^{\ast}, \Lambda_t^{\ast})$ admits a rank-$K_t$ factorization:
\begin{equation*}
    Q_t^{\ast} = W_t^{\ast} P_t^{\ast\top}, \qquad
    \Lambda_t^{\ast} = W_t^{\ast} A_t^{\ast\top}, \qquad
    |\ci_t(W_t^{\ast})| = K_t,
\end{equation*}
where $W_t^{\ast} \in [0,1]^{n_t \times K}$ represents document–topic mixing weights and satisfies $\sum_k w_{tik}^{\ast}=1$ for each document, $\ci_t(W_t^{\ast}) = \{k\in[K]: \|W_{tk}^{\ast}\|_1>0\}$ denotes the active topic indices, $P_t^{\ast} \in (0,1)^{p\times K}$ encodes topic-specific word occurrence probabilities, and $A_t^{\ast} \in \mathbb{R}_+^{p\times K}$ encodes topic-specific repetition intensities.

Since activation and repetition are modeled as two signals, this factorization separates whether a word appears from how strongly it repeats once present. Both $P_t^{\ast}$ and $A_t^{\ast}$ are governed by the same document-level mixture, reflecting the intuition that occurrence patterns and repetition intensities shift coherently across topics.

Due to the latent low-rank structure and temporal evolution, the model is identifiable only under additional structural conditions.
First, the factorization admits intrinsic label-switching ambiguity and remains invariant under a common column permutation. 
Second, if topic $k$ is inactive at time $t$, the corresponding column of $W_t^{\ast}$ is zero, and the associated columns of $P_t^{\ast}$ and $A_t^{\ast}$ are not identifiable from the likelihood.

To remove this ambiguity, we align topic labels across time by minimizing a temporal deviation criterion. Let $\tau_P,\tau_A>0$ be fixed constants. Define
\begin{equation}\label{Equ:tilde_g}
    \tilde{g}(\{(W_t^{\ast}, P_t^{\ast}, A_t^{\ast})\}_{t=1}^{T})
 := \tau_P \sum_{t=1}^{T} \sum_{k\in\ci_t} 
    \|P_{tk}^{\ast} - \tilde{P}_{tk}^{\ast}\|_2^2
  + \tau_A \sum_{t=1}^{T} \sum_{k\in\ci_t} 
    \|A_{tk}^{\ast} - \tilde{A}_{tk}^{\ast}\|_2^2,
\end{equation}
where $\tilde P_{tk}^{\ast}$ and $\tilde A_{tk}^{\ast}$ denote temporal averages over windows in which topic $k$ is active, that is
\begin{equation*}
    \tilde{P}_{tk}^{\ast} = \frac{\sum_{t': k\in\ci_{t'}(W_{t'}^{\ast})} P_{t'k}^{\ast}}{|\{t':k\in\ci_{t'}(W_{t'}^{\ast})\}|},
    \qquad
    \tilde{A}_{tk}^{\ast} = \frac{\sum_{t': k\in\ci_{t'}(W_{t'}^{\ast})} A_{t'k}^{\ast}}{|\{t':k\in\ci_{t'}(W_{t'}^{\ast})\}|}.
\end{equation*}
We search over permutation matrices $R_t \in \cR^{K\times K}$ to minimize 
$\tilde g(\{(W_t^{\ast},P_t^{\ast},A_t^{\ast})R_t\}_{t=1}^T)$. Without loss of generality, we assume the true parameters have been aligned in this way, and inactive topic–word profiles are fixed to their temporal averages.

Let $\theta_t = (W_t, P_t, A_t)$ denote the tuple of parameter matrices at time $t$, and let $\theta = (\theta_1,\cdots,\theta_T)$ collect all parameters across time. We use $\theta_t^{\ast}$ and $\theta^{\ast}$ to denote the corresponding true parameters. 
We now define the feasible parameter space and state identifiability conditions.

\begin{defi}[Feasible region]\label{Def:feasible}
For fixed $K$, we define the feasible region $\cc_t := \cc_{W_t} \times \cc_P \times \cc_A$, where
\BEqn
& & \cc_{W_t} = \{W_t\in[0,1]^{n_t\times K}: ~\sum_k w_{tik}=1 ~\mbox{for } i=1,\cdots,n_t\}, \\ 
& & \cc_{P} = \{P\in[l_p,u_p]^{p\times K}\}, \qquad
    \cc_{A} = \{A\in[l_a,u_a]^{p\times K}: ~\sum\nolimits_{j=1}^{p} a_{jk}=p ~\mbox{for }k=1,\cdots,K\},
\EEqn
with constants $0<l_p<u_p<1$ and $0<l_a<1<u_a$ independent of $\{n_t\}_{t=1}^{T}$ and $p$. 
\end{defi}

The bounds in $\cc_P$ and $\cc_A$ keep the log-likelihood well-defined and ensure sufficient local curvature. Such boundedness conditions are standard in the analyses of graph and hypergraph models \citep{ma2020universal,wu2024general}. 
The column-sum constraint in $\cc_A$ fixes the scaling of $\Lambda_t$ and parallels the normalizations used in pLSI-type models. We use a column sum of $p$ to keep $A_t$ on a comparable scale to $P_t$, which simplifies both optimization and theory.

\begin{rmk}
At the cost of additional technical arguments, we may allow the bounds $l_p$ and $1-u_p$ to decay with $p$, provided that all but $O(1)$ entries remain bounded away from $0$ and $1$, i.e., $\#\{p_{tjk}:~p_{tjk}<c_0 \mbox{ or } p_{tjk}>1-c_0\} = O(1)$ for some fixed constant $c_0$ independent of $p$. For clarity of the theory, we work with constant bounds independent of $p$ throughout this article.
\end{rmk}

We impose identifiability conditions ensuring that $\theta^{\ast}$ is unique up to a global topic permutation; a detailed formulation is given in Appendix \ref{Sec:Full_Assumptions}.

\begin{assumpt}[Identifiability (informal)]\label{Assumpt:identifiability_main}
We assume that:
\begin{enumerate}[label=(\roman*)]
    \item \emph{(Feasibility)} The true parameters lie in the feasible region, and $|\ci_t(W_t^{\ast})| = K_t$.

    \item \emph{(Within-time identifiability)}         
    For each $t$, every active topic admits an anchor document, and the active joint topic–word vectors $\{ z_{tk}^{\ast} = (P_{tk}^{\ast\top}, A_{tk}^{\ast\top}) \}_{k \in \ci_t(W_t^{\ast})}$ are linearly independent and satisfy a standard extreme-point condition.

    \item \emph{(Cross-time identifiability)} 
    Each topic is active in at least one time window, the graph on time indices induced by shared active topics is connected, and the temporal deviation criterion admits a unique minimizer up to a common permutation.
\end{enumerate}
\end{assumpt}

Assumption \ref{Assumpt:identifiability_main} formalizes correct model specification and ensures identifiability both within and across time. 
Within each time window, the anchor-document and separability conditions guarantee uniqueness of the bilinear factorization up to permutation. Anchor-type conditions are standard in mixed-membership models; see \cite{klopp2023assigning, jung2024graph}. Anchor-word assumptions used in pLSI methods \citep{arora2012learning,arora2013practical,ke2024using} are sufficient and typically stronger than our separability condition.
Across time, the coverage and connectivity condition allows topics to appear or disappear while preserving global identifiability, and the unique-alignment condition rules out degenerate cases in which multiple alignment sequences yield identical temporal deviation.

Under these conditions, the low-rank factorization $\theta^{\ast}$ is identifiable up to permutation.

\begin{prop}[Uniqueness of true parameters]\label{Prop:uniqueness}
Define the equivalence class
\begin{equation*}
    \Theta
 := \lrcp{\theta=(\theta_1,\cdots,\theta_T): 
          ~\theta_t=(W_t, P_t, A_t)\in\cc_t, 
          ~|\ci_t(W_t)| = K_t,
          ~Q_t^{\ast} = W_t P_t^{\top}, 
           \Lambda_t^{\ast} = W_t A_t^{\top} ~\forall t}.
\end{equation*}
Fix $\tau_P, \tau_A > 0$ and define
\begin{equation*}
    g(\theta)
  = \tau_P \sum\limits_{t=1}^{T} 
    \|P_t - T^{-1} \sum_{t'=1}^{T} P_{t'}\|_F^2
  + \tau_A \sum\limits_{t=1}^{T} 
    \|A_t - T^{-1} \sum_{t'=1}^{T} A_{t'}\|_F^2.
\end{equation*}
Under Assumption \ref{Assumpt:identifiability_main}, for any $\tilde\theta \in \argmin_{\theta \in \Theta} g(\theta)$, there exists a permutation matrix $R \in \cR^{K\times K}$ such that for all $t$, $W_t^{\ast} = \tilde{W}_t R$, $P_t^{\ast} = \tilde{P}_t R$ and $A_t^{\ast} = \tilde{A}_t R$.
Consequently, $\theta^{\ast}$ is the unique minimizer of $g(\theta)$ over $\Theta$ up to a common column permutation.
\end{prop}

Proposition \ref{Prop:uniqueness} provides a well-defined target (up to permutation) for non-asymptotic error analysis in Section \ref{Sec:Theory}.

\subsection{Estimation}\label{Sec:Method_Estimation}

We estimate the model parameters by maximizing the H-Multinomial likelihood given by Definition \ref{Def:H_Multinomial} over the feasible region with a direct temporal regularization imposed on the topic-word profiles to preserve temporal consistency in topic semantics. Since documents are independently generated over time, no temporal structure is imposed on the document–topic proportions $\{W_t\}_{t=1}^{T}$.

Throughout, we denote $Q_t = W_t P_t^{\top}$ and $\Lambda_t = W_t A_t^{\top}$ for each $t$. We decompose the empirical negative log-likelihood into contributions from the Bernoulli and multinomial components of the H-Multinomial distribution, that is
\begin{eqnarray*}
    \hat\ell_{t,i}^{(1)}(\theta)
&=& - \sum\limits_{j=1}^{p} \lrcp{e_{tij}\log(q_{tij}) + (1-e_{tij})\log(1-q_{tij})}, \\
    \hat\ell_{t,i}^{(2)}(\theta)
&=& - \sum\limits_{j=1}^{p} \lrcp{r_{tij} \log\frac{\lambda_{tij}}{\sum_u \lambda_{tiu}e_{tiu}}},
\end{eqnarray*}
and let $\hat\ell_t^{(v)}=\sum_{i=1}^{n_t}\hat\ell_{t,i}^{(v)}$ and
$\hat\ell^{(v)}=\sum_{t=1}^{T}\hat\ell_t^{(v)}$ for $v=1,2$. Here, $q_{tij} = \bigag{w_{ti}, p_{tj}}$ and $\lambda_{tij} = \bigag{w_{ti}, a_{tj}}$ with $w_{ti}$ denoting the $i$-th row of $W_t$ and $p_{tj}$ and $a_{tj}$ denoting the $j$-th rows of $P_t$ and $A_t$, respectively. Here, we omit additive terms depending only on $s_{ti}$ and $\{r_{tij}\}_{j=1}^{p}$, as they do not affect the optimization.

Because the low-rank factorization is permutation invariant, topic labels must be aligned across time prior to imposing temporal regularization. Specifically, we set $O_1=I_K$ and determine $O_t$ for $t\ge2$ by
\begin{equation}\label{Equ:Alignment}
    O_t
  = \argmin_{O\in\cR^{K\times K}}
    \lrcp{
      \tau_{P} \|P_tO - P_{t-1}O_{t-1}\|_F^2
    + \tau_{A} \|A_tO - A_{t-1}O_{t-1}\|_F^2}, ~t\geq 2
\end{equation}
where $\tau_P, \tau_A > 0$ are tuning parameters. In practice, this assignment problem can be efficiently solved using the Hungarian algorithm.
\begin{rmk}
The permutation updates affect only the temporal regularization terms and do not alter the likelihood value by noting that $Q_t = W_t P_t^{\top} = (W_t O_t) (P_t O_t)^{\top}$ and $\Lambda_t = W_t A_t^{\top} = (W_t O_t) (A_t O_t)^{\top}$. 
This sequential alignment provides a computationally tractable approximation to the global alignment criterion in \eqref{Equ:tilde_g}, while preserving temporal consistency.
\end{rmk}

We estimate the parameters by solving the penalized optimization problem
\begin{equation}\label{Equ:Optimization}
    \hat\theta
  = \argmin_{\stackrel{\theta=(\theta_1,\cdots,\theta_T)}{\theta_t=(W_t,P_t,A_t) \in \cc_t}} f(\theta),
    \qquad
    f(\theta)
  = \hat\ell^{(1)}(\theta) + \hat\ell^{(2)}(\theta)
  + \tau_P g_P(\theta) + \tau_A g_A(\theta),
\end{equation}
where the temporal regularization is defined as
\begin{equation*}
    g_P(\theta) 
  = \sum\limits_{t=1}^{T} \|P_tO_t - \frac{1}{T}\sum_{t'=1}^{T} P_{t'}O_{t'}\|_F^2, \qquad
    g_A(\theta) 
  = \sum\limits_{t=1}^{T} \|A_tO_t - \frac{1}{T}\sum_{t'=1}^{T} A_{t'}O_{t'}\|_F^2,
\end{equation*}
In implementation, alignment and parameter updates are performed alternately: given current estimates, we update $\{O_t\}_{t=1}^{T}$ via \eqref{Equ:Alignment}, and then minimize $f(\theta)$ with fixed permutations. 

Over the feasible region defined as Definition \ref{Def:feasible}, all logarithmic terms in $f(\theta)$ are well-defined. The objective $f(\theta)$ combines the Bernoulli and multinomial components of the H-Multinomial likelihood with temporal regularization on topic–word parameters $(P_t,A_t)$. This formulation enforces smooth semantic evolution while allowing document–topic proportions $W_t$ to vary freely across time. When $T=1$, the regularization terms vanish, and the problem reduces to the static estimation setting.

The separation of occurrence and repetition enables direct temporal regularization on the topic-specific semantic signals $(P_t, A_t)$. This yields a geometrically interpretable decomposition as $W_t$ tracks the temporal document-topic prevalence, whereas $P_t$ and $A_t$ characterize the evolution of topic-word behaviors.

Unlike dynamic LDA-type models that impose state-space evolution on latent variables, our formulation regularizes topic-specific word distributions directly, permitting localized semantic shifts while maintaining global coherence. In contrast to spectral pLSI approaches \citep{arora2012learning, arora2013practical, klopp2023assigning, ke2024using} relying on time-specific anchor geometry without an explicit temporal association, our likelihood-based formulation accommodates time-varying corpus geometry and remains stable under topic appearance or disappearance.

\subsection{Projected Gradient Descent Algorithm}\label{Sec:Method_Algorithm}

The objective function $f(\theta)$ is nonconvex due to the bilinear factorization and the support-dependent normalization in the multinomial component. Closed-form solutions are unavailable, and global minimizers need not be unique. We therefore adopt projected gradient descent (PGD), summarized in Algorithm \ref{Alg:PGD}.

\begin{algorithm}[h!]
\caption{Projected Gradient Descent for solving optimization problem (\ref{Equ:Optimization})}\label{Alg:PGD}
\setlength{\baselineskip}{.65\baselineskip}
\begin{algorithmic}[1]
\Require Topic number $K$, (misaligned) initialization $\tilde\theta^{(0)}$, step sizes $\eta_{W_t},\eta_{P_t},\eta_{A_t}>0$, tuning parameters $\tau_P,\tau_A>0$, maximum iterations $L$, tolerance $\varepsilon>0$.

\vspace{3mm}
\State Set $O_1^{(0)}=I_K$ and compute $O_t^{(0)}$ using Equation (\ref{Equ:Alignment}) for $t=2,\cdots,T$.
\Comment{Alignment}
\vspace{1mm}
\State Align $(\hat{W}_t^{(0)}, \hat{P}_t^{(0)}, \hat{A}_t^{(0)}) \leftarrow (\tilde{W}_t^{(0)}, \tilde{P}_t^{(0)}, \tilde{A}_t^{(0)})O_t^{(0)}$ for $t=2,\cdots,T$.
\vspace{3mm}

\For{$\ell = 0, 1, \cdots, L$} 
    \State Compute temporal averages $\bar{P}^{(\ell)} = \frac{1}{T}\sum_{t=1}^{T} \hat{P}_t^{(\ell)}$ and $\bar{A}^{(\ell)} = \frac{1}{T}\sum_{t=1}^{T} \hat{A}_t^{(\ell)}$.
    
    \For{$t=1,\cdots, T$}
        \State Compute $Q_t^{(\ell)} = \hat{W}_t^{(\ell)} \hat{P}_t^{(\ell)\top}$ and $\Lambda_t^{(\ell)} = \hat{W}_t^{(\ell)} \hat{A}_t^{(\ell)\top}$. 
        \vspace{2mm}

        \State Compute gradient matrices $\Sigma_{Q_t}^{(\ell)}, \Sigma_{\Lambda_t}^{(\ell)} \in \br^{n_t\times p}$ with
        \begin{equation*}
            (\Sigma_{Q_t}^{(\ell)})_{ij} = \frac{q_{tij}^{(\ell)}-e_{tij}}{q_{tij}^{(\ell)}(1-q_{tij}^{(\ell)})}, \qquad
            \displaystyle{(\Sigma_{\Lambda_t}^{(\ell)})_{ij} = \frac{s_{ti} e_{tij}}{\sum_u \lambda_{tiu}^{(\ell)} e_{tiu}} - \frac{r_{tij}}{\lambda_{tij}^{(\ell)}}}
        \end{equation*}        
        \vspace{2mm}
        
        \State $\tilde{W}_t^{(\ell+1)} \leftarrow \Pi_{\cc_{W_t}}\lrp{\hat{W}_t^{(\ell)} - \eta_{W_t} (\Sigma_{Q_t}^{(\ell)} \hat{P}_t^{(\ell)} + \Sigma_{\Lambda_t}^{(\ell)} \hat{A}_t^{(\ell)}) }$.      
        \Comment{Iterative update}
        \vspace{1mm}

        \State $\tilde{P}_t^{(\ell+1)} \leftarrow \Pi_{\cc_P}\lrp{\hat{P}_t^{(\ell)} - \eta_{P_t} (\Sigma_{Q_t}^{(\ell)\top} \hat{W}_t^{(\ell)} + \tau_P (\hat{P}_t^{(\ell)} - \bar{P}^{(\ell)}))}$.    
        \vspace{1mm}
        
        \State $\tilde{A}_t^{(\ell+1)} \leftarrow \Pi_{\cc_A}\lrp{\hat{A}_t^{(\ell)} - \eta_{A_t} (\Sigma_{\Lambda_t}^{(\ell)\top} \hat{W}_t^{(\ell)} + \tau_A (\hat{A}_t^{(\ell)} - \bar{A}^{(\ell)})) }$.    
        \vspace{1mm}

        \If{$t \geq 2$}
        \Comment{Alignment per update}
            \State Update permutation $O_t^{(\ell+1)}$ via Equation (\ref{Equ:Alignment}). 
            \vspace{2mm}
            \State Realign 
            $(\hat{W}_t^{(\ell+1)}, \hat{P}_t^{(\ell+1)}, \hat{A}_t^{(\ell+1)}) \leftarrow (\tilde{W}_t^{(\ell+1)}, \tilde{P}_t^{(\ell+1)}, \tilde{A}_t^{(\ell+1)}) O_t^{(\ell+1)}$.
        \EndIf
    \EndFor    

    \State Compute $\delta^{(\ell)} = \max_t \{\|\hat{W}_t^{(\ell+1)} - \hat{W}_t^{(\ell)}\|_F, \|\hat{P}_t^{(\ell+1)} - \hat{P}_t^{(\ell)}\|_F, \|\hat{A}_t^{(\ell+1)} - \hat{A}_t^{(\ell)}\|_F\}$.
    \vspace{1mm}
    
    \If{$\delta^{(\ell)} < \varepsilon$}    
    \Comment{Early stopping}
        \State \textbf{break}
    \EndIf    
\EndFor
\vspace{3mm}

\Ensure $\hat{W}_t = \hat{W}_t^{(\ell+1)}, \hat{P}_t = \hat{P}_t^{(\ell+1)}, \hat{A}_t = \hat{A}_t^{(\ell+1)}$.
\end{algorithmic}
\end{algorithm}

The projection operator $\Pi_{\cc}$ denotes the Euclidean projection onto the feasible set. For $P_t$, this reduces to elementwise truncation onto $[l_p,u_p]$. For $W_t$, it amounts to rowwise projection onto the probability simplex in $\mathbb{R}^K$, which can be implemented efficiently using sorting-based methods or Michelot’s algorithm. For $A_t$, the projection reduces to the columnwise projection onto a capped simplex with fixed sum, which is a natural generalization of the simplex projection used for $W_t$.

After each iteration $\ell$, we update the column permutation matrices to align the estimates $(\hat{W}_t^{(\ell)}, \hat{P}_t^{(\ell)}, \hat{A}_t^{(\ell)})$ across time. The permutation updates are treated as part of the algorithmic procedure rather than the optimization variables. Since $Q_t$ and $\Lambda_t$ are permutation-invariant, these updates affect the objective function only through the temporal regularization terms and do not change the likelihood value.

\begin{rmk}
Our theory assumes an initialization in a local neighborhood of $\theta^{\ast}$. The heuristic initializer used in practice is given in Appendix \ref{Sec:Method_Init}.
\end{rmk}

\section{Theoretical Properties}\label{Sec:Theory}

In this section, we establish local convergence guarantees and finite-sample error bounds for the proposed estimation algorithm.  
Section \ref{Sec:Theory_Regularity} introduces the error metric and basic regularity conditions.
Section \ref{Sec:Theory_Main} states the main theoretical results, both in deterministic and probabilistic forms.
Section \ref{Sec:Theory_Est_K} provides a consistent estimation of the topic number $K$.

\subsection{Regularity Assumptions}\label{Sec:Theory_Regularity}

To ensure topic separability, we impose a standard non-degeneracy condition requiring the true low-rank factors to be well-conditioned.

\begin{assumpt}[Non-degeneracy]\label{Assumpt:NonDegeneracy}
Let $\sigma_k(\cdot)$ denote the $k$-th largest singular value. For each $t=1,\cdots,T$, we assume that
\BEqn
& & \underline{c}_{W_t^{\ast}} \sqrt{n_t}
\le \sigma_{\min}(W_{t,\ci_t(W_t^{\ast})}^{\ast})
\le \sigma_{\max}(W_t^{\ast})
\le \bar{c}_{W_t^{\ast}} \sqrt{n_t}, \\
& & \underline{c}_{P_t^{\ast}} \sqrt{p}
\le \sigma_{\min}(P_t^{\ast})
\le \sigma_{\max}(P_t^{\ast})
\le \bar{c}_{P_t^{\ast}} \sqrt{p}, \\
& & \underline{c}_{A_t^{\ast}} \sqrt{p}
\le \sigma_{\min}(A_t^{\ast})
\le \sigma_{\max}(A_t^{\ast})
\le \bar{c}_{A_t^{\ast}} \sqrt{p},
\EEqn
where all constants are positive and independent of $\{n_t\}_{t=1}^{T}$ and $p$.
Here, $W_{t,\ci_t(W_t^{\ast})}^{\ast}$ denotes the sub-matrix of $W_t^{\ast}$ formed by its active (nonzero) columns, with $\ci_t(W_t^{\ast}) = \{k\in[K]: \|W_{tk}^{\ast}\|_1 > 0\}$.
\end{assumpt}

Assumption \ref{Assumpt:NonDegeneracy} excludes pathological cases where topics become nearly collinear or vanish over time, which is standard in analyses of low-rank factor models and spectral topic methods \citep{arora2012learning, ke2024using, klopp2023assigning, fan2025covariates}. This assumption is natural for $P_t^{\ast},A_t^{\ast}$ with bounded entries and mild for $W_t^{\ast}$ when topic prevalence is comparable.

Due to the intrinsic permutation invariance of topic models, estimation error must be measured after appropriate label alignment. For $\theta = (\theta_1,\cdots,\theta_T)$ with $\theta_t=(W_t, P_t, A_t)$, we define the oracle-aligned error metric:
\begin{equation}\label{Equ:Error}
    e(\theta)
  = \sum\limits_{t=1}^{T}  
    \min\limits_{R_t\in\cR^{K\times K}}
    \lrcp{  \kappa_{W_t} \|W_t R_t - W_t^{\ast}\|_F^2 
          + \kappa_{P_t} \|P_t R_t - P_t^{\ast}\|_F^2
          + \kappa_{A_t} \|A_t R_t - A_t^{\ast}\|_F^2 },
\end{equation}
where $\{R_t\in\cR^{K\times K}\}_{t=1}^{T}$ are permutation matrices that jointly attain the minimum of $e(\theta)$. For notational convenience, we write $e^{(\ell)} = e(\hat\theta^{(\ell)})$. We introduce block-specific weights proportional to the operator norms of the true factors, that is
\begin{equation*}
    \kappa_{W_t} = \|P_t^{\ast}\|_{op}^2 + \|A_t^{\ast}\|_{op}^2,
    \qquad
    \kappa_{P_t} = \|W_t^{\ast}\|_{op}^2,
    \qquad
    \kappa_{A_t} = \|W_t^{\ast}\|_{op}^2.
\end{equation*}
With this choice, each block in Equation \eqref{Equ:Error} is measured on a balanced scale, and the resulting metric is tailored to the local curvature of the objective function.

Due to the nonconvexity of the objective and the nonuniqueness of the global optimizer, our analysis focuses on the local behavior of the projected gradient descent (PGD) iterates within a neighborhood of the truth. We therefore define the local basin as follows.

\begin{defi}[Local basin]\label{Def:basin}
For $r>0$, define the basin of radius $r>0$ around $\theta^{\ast}$ by
\begin{equation*}
    \cb_r(\theta^{\ast}) 
  = \{\theta=(\theta_1,\cdots,\theta_T): ~\theta_t = (W_t,P_t,A_t) \in \cc_t, ~e(\theta) \le r\}.
\end{equation*}
\end{defi}

Definition \ref{Def:basin} translates geometric contraction into oracle-aligned error control. Establishing finite-sample bounds requires characterizing the local curvature of the empirical objective, which is nonstandard due to support-dependent normalization, componentwise dependence in the repetition-layer, Bernoulli masking, and stochastic perturbations arising from sampling noise. We therefore impose the following local regularity conditions along the PGD trajectory.

\begin{assumpt}[Local regularity (informal)]\label{Assumpt:Local_short}
Along the PGD trajectory within the local basin $\cb_r(\theta^{\ast})$, we assume that the empirical objective satisfies: (i) a local restricted strong convexity (RSC) property in the bilinear directions associated with $(W_t,P_t)$ and $(W_t,A_t)$; (ii) a non-alignment condition preventing first-order perturbation terms from concentrating along the support-normalization direction; (iii) uniform self-normalized concentration of the repetition noise along the interpolation path; (iv) an energy-capture condition ensuring that the Bernoulli sampling mask preserves the dominant error directions; and (v) strictly positive lower bound on the effective curvature of the multinomial component.
\end{assumpt}

The formal statements are provided in the supplement; see Appendix \ref{Sec:Full_Assumptions}. These conditions ensure stable local curvature and control stochastic fluctuations, which are analogous to local RSC and concentration assumptions in nonconvex low-rank estimation.

\subsection{Initialization and Main Results}\label{Sec:Theory_Main}

To analyze Algorithm \ref{Alg:PGD}, we require the initialization to lie within a local basin of $\theta^{\ast}$, which is commonly employed in nonconvex optimizations \citep{ma2018implicit, chen2018projected, chi2019nonconvex, lyu2023optimal, zhang2023generalized}. The admissible basin radius depends jointly on cross-time topic separability and within-time topic discrepancy.

Specifically, for the $(P_t^{\ast},A_t^{\ast})$ blocks, we denote by $b^{\ast}$ the minimum permutation-induced temporal separation gap, by $B^{\ast}$ the maximum temporal discrepancy, by $c_{PA}^{\ast}$ the minimum within-time cross-topic discrepancy, and by $C_{PA}^{\ast}$ its maximum counterpart. For the document–topic matrices, we denote by $C_W^{\ast}$ the maximal within-time cross-topic discrepancy. Detailed formulations are given in Appendix \ref{Sec:Full_Init}. Collectively, these quantities quantify the separability of topic–word profiles and topic prevalence across all time windows.

\begin{assumpt}[Initialization]\label{Assumpt:Init}
Let $\hat\theta^{(0)}$ denote the initializer of Algorithm~\ref{Alg:PGD}. We assume that $\hat\theta^{(0)} \in \cb_{r_0}(\theta^{\ast})$, where $r_0>0$ satisfies
\begin{equation}\label{Equ:Init}
    r_0 
  < \min\left\{
        \left(\sqrt{4B^{\ast}+\frac{1}{2}b^{\ast}}-2\sqrt{B^{\ast}}\right)^2,
        \frac{(c_{PA}^{\ast} - C_W^{\ast})^2}{4(C_W^{\ast} + C_{PA}^{\ast})}
      \right\}.
\end{equation}
\end{assumpt}

The first term in \eqref{Equ:Init} prevents incorrect temporal alignments of the $(P_t,A_t)$ blocks from becoming locally optimal, while the second term ensures that within-time topic permutations cannot reduce the objective. Together, these conditions guarantee that the algorithmic alignment $\{O_t\}_{t=1}^{T}$ coincides with the oracle alignment $\{R_t\}_{t=1}^{T}$ throughout the iterations. 

Note that the right-hand side of \eqref{Equ:Init} scales on the order of $\min_t\{n_t p\}$, then the basin condition is satisfied if the entrywise initialization error of each latent factor is uniformly bounded by a sufficiently small constant. Such radius requirements are standard in local convergence analyses of nonconvex models; see, for example, \cite{chen2015fast, ma2020universal, zhang2023generalized}.

\begin{thm}[Deterministic local convergence]\label{Thm:Deterministic}
Suppose that Assumptions \ref{Assumpt:rho}–\ref{Assumpt:identifiability_main} and Assumptions \ref{Assumpt:NonDegeneracy}–\ref{Assumpt:Init} hold. We choose $\tau_P = \tau_A > 0$.
Let $\{\hat\theta^{(\ell)}\}_{\ell\ge0}$ denote the PGD trajectory starting from 
$\hat\theta^{(0)} \in \cb_{r_0}(\theta^{\ast})$, with blockwise step sizes $\eta_{W_t} = \kappa_{W_t}^{-1} \eta_0$, $\eta_{P_t} = \kappa_{P_t}^{-1} \eta_0$ and $\eta_{A_t} = \kappa_{A_t}^{-1} \eta_0$.
Define
\BEqn
    \cn
&:=& \sum\nolimits_{t=1}^{T}
    \lrp{  \|  \Sigma_{Q_t}(\theta^{\ast})
             - \Sigma_{Q_t}^{\ast}(\theta^{\ast})\|_{op}^2 
         + \|  \Sigma_{\Lambda_t}(\theta^{\ast})
             - \Sigma_{\Lambda_t}^{\ast}(\theta^{\ast})\|_{op}^2 }, \\
    \ct_P
&:=& \sum\nolimits_{t=1}^{T} 
    \|W_t^{\ast}\|_{op}^{-2} \|P_t^{\ast} - \bar{P}^{\ast}\|_F^2, 
    \qquad
    \ct_A
 := \sum\nolimits_{t=1}^{T} 
    \|W_t^{\ast}\|_{op}^{-2} \|A_t^{\ast} - \bar{A}^{\ast}\|_F^2.
\EEqn
and let $\ci = c_{\Sigma}\cn + c_{P}\ct_P + c_{A}\ct_A$, where $\Sigma_{Q_t}, \Sigma_{\Lambda_t}$ denote the gradient with respect to $Q_t$ and $\Lambda_t$ of the empirical loss, and $\Sigma_{Q_t}^{\ast}, \Sigma_{\Lambda_t}^{\ast}$ denote the counterparts of the population loss.

There exist constants $\tilde{r}_0^{\dagger}, \psi_0 > 0$ independent of 
$\{n_t\}_{t=1}^{T}$ and $p$, and a step-size upper bound 
$\eta_0^{\dagger}(\{n_t\}_{t=1}^{T}, p, \bar{R}) > 0$, such that if
\begin{equation*}
    0 
  < r_0 
  < \tilde{r}_0^{\dagger}
    \min_{1\le t\le T}
    \{ \|W_t^{\ast}\|_{op}^2 \|P_t^{\ast}\|_{op}^2, 
      ~\|W_t^{\ast}\|_{op}^2 \|A_t^{\ast}\|_{op}^2\}, 
    \quad
    0 
  < \eta_0 
  < \eta_0^{\dagger}(\{n_t\}_{t=1}^{T}, p, \bar{R}),   
    \quad
    \ci < \psi_0 r_0.
\end{equation*}
then for all $\ell \ge 0$, $\hat\theta^{(\ell)} \in \cb_{r_0}(\theta^{\ast})$ and
\begin{equation*}
    e^{(\ell)}
\le (1 - \eta_0 \psi_0)^{\ell} e^{(0)}
  + \psi_0^{-1} \lrp{1 - (1 - \eta_0 \psi_0)^{\ell}} \ci.
\end{equation*}
\end{thm}

Theorem \ref{Thm:Deterministic} establishes linear contraction of PGD at rate $1-\eta_0\psi_0$ toward a deterministic neighborhood of radius $\ci/\psi_0$ around the truth. The perturbation term $\ci$ captures curvature fluctuations from the Bernoulli–multinomial components and temporal drift of topic–word profiles. Linear contraction holds when these perturbations remain small relative to the local curvature $\psi_0$.

By establishing concentration bounds for the perturbation term $\cn$, we translate the deterministic contraction result of Theorem \ref{Thm:Deterministic} into an explicit high-probability guarantee.

\begin{thm}[Probabilistic local convergence]\label{Thm:Probabilistic}
Suppose the conditions of Theorem \ref{Thm:Deterministic} hold and assume that there exists a constant $c_n > 0$ such that $\min_{1\le t\le T} \{n_t/n\} \geq c_n$.
For each $\delta\in(0,1)$, there exists constants $\psi_0, \psi_1, \psi_2 > 0$ such that if $0 < r_0 < r_0^{\dagger}$, $0 < \eta_0 < \underline{\eta}_0^{\dagger}(\{n_t\}_{t=1}^{T},p;\delta)$, and 
\begin{equation}\label{Equ:Prob_Regime}
    \psi_1 \max\{n, p\} \log^2\lrp{\frac{n+p}{\delta}}
\le \psi_0 \psi_2 np \cdot \frac{r_0}{r_0^{\dagger}},
\end{equation}
then with probability at least $1-\delta$, for all $\ell\geq0$, it holds that $\hat\theta^{(\ell)} \in \cb_{r_0}(\theta^{\ast})$ and
\begin{equation*}
    e^{(\ell)}
    \le (1-\eta_0\psi_0)^{\ell} e^{(0)}
    + \psi_0^{-1} \psi_1 \lrp{1 - (1 - \eta_0 \psi_0)^{\ell}}
      \max\{n, p\} \log^2\lrp{\frac{n+p}{\delta}},
\end{equation*}
\end{thm}

Theorem \ref{Thm:Probabilistic} shows that $\ci$ is of order $\max\{n,p\}\log^2((n+p)/\delta)$ with high probability. In contrast to existing spectral pLSI results (e.g., \cite{klopp2023assigning}), where total repetition counts are treated as fixed, our model introduces additional randomness through the Bernoulli sampling mechanism under Assumption \ref{Assumpt:rho}. Furthermore, the support-dependent normalization couples repetition noise across activated nodes and induces nonlinear dependence in the compositional parameters. Consequently, the dependence structure is more intricate than standard multinomial models, making the multinomial perturbation arguments inapplicable. To address this, we develop a novel conditioning technique that enables a matrix Bernstein-type concentration inequality tailored to the hypergraph-induced likelihood.

The signal-to-noise condition \eqref{Equ:Prob_Regime} ensures that the stochastic perturbation remains sufficiently small relative to the effective local curvature. When the sample sizes $\{n_t\}_{t=1}^{T}$ are balanced across time, and $n, p$ grow at most polynomially relative to each other, this condition holds for sufficiently large $n$ or $p$.

As a direct consequence, we derive the Frobenius-norm bounds for each block.

\begin{cor}\label{Cor:ErrorBound_WPA}
Under the conditions of Theorem~\ref{Thm:Probabilistic}, for any $\delta\in(0,1)$, there exists deterministic constants $c_W, c_P, c_A$, such that with probability at least $1-\delta$, it holds for each $t=1,\cdots,T$ that
\BEqn
    n_t^{-1/2} \|\hat{W}_t R_t - W_t^{\ast}\|_F
&\le& c_W \lrp{\frac{\max\{n,p\}}{n_t p}}^{1/2}
    \log\lrp{\frac{n+p}{\delta}}, \\
    p^{-1/2} \|\hat{P}_t R_t - P_t^{\ast}\|_F
&\le& c_P \lrp{\frac{\max\{n,p\}}{n_t p}}^{1/2}
    \log\lrp{\frac{n+p}{\delta}}, \\
    p^{-1/2} \|\hat{A}_t R_t - A_t^{\ast}\|_F
&\le& c_A \lrp{\frac{\max\{n,p\}}{n_t p}}^{1/2}
    \log\lrp{\frac{n+p}{\delta}},
\EEqn
where $\hat{W}_t$ denotes the limit point of $\hat{W}_t^{(\ell)}$ and analogous for $\hat{P}, \hat{A}$.
\end{cor}

Under balanced growth of $n$ and $p$, Corollary \ref{Cor:ErrorBound_WPA} yields rate $O\lrp{\log(n+p)/\sqrt{\min(n,p)}}$. Compared to static spectral analyses \citep{klopp2023assigning, ke2024using}, our result accommodates temporal evolution, avoids geometric growth constraints between $n$ and $p$, and provides simultaneous guarantees for all latent blocks $(W_t,P_t,A_t)$. Specifically, the rate in \cite{klopp2023assigning} scales as $O(\sqrt{\log(n+p)/p})$ under their document-length regime, which can be faster when $p \gg n$, but their analysis only provides guarantees for the document-topic matrix. The rate in \cite{ke2024using} scales as $O\lrp{\sqrt{p\log(n+p)/n} + \sqrt{\log(n+p)/p}}$ and becomes sharper when $n \gg p^2$, whereas in moderate or balanced regimes our bound is tighter.

\subsection{Consistent Estimation of $K$}\label{Sec:Theory_Est_K}

So far, all analysis has been conducted under the assumption that the true topic number $K$ is known. In practice, $K$ is typically unknown and must be estimated. When temporal drift is mild, $K$ equals the common rank of the leading signal component of ${Q_t^{\ast}}$ and ${\Lambda_t^{\ast}}$. Since $\Lambda_t^{\ast}$ is not directly observable due to support-dependent normalization, we estimate $K$ from the Bernoulli component.

Specifically, we stack $\{E_t\}_{t=1}^{T}$ into a matrix $E = (E_1^{\top}, \cdots, E_T^{\top})^{\top} \in \{0,1\}^{n\times p}$ with $n=\sum_{t=1}^{T} n_t$. Under Assumption \ref{Assumpt:identifiability_main}, the expected value of $E$ admits a rank-$K$ leading component plus temporal deviation and sampling noise. When the sample sizes are sufficiently large, the leading $K$ singular value of $ E$ is separated from the remaining spectrum with high probability. We therefore estimate $\hat{K} := \# \{k\geq1: \sigma_k(E) > \tau_{n,p}\}$. The following theorem establishes consistency.

\begin{thm}[Consistency of $\hat{K}$]\label{Thm:RankConsistency}
Suppose that Assumption \ref{Assumpt:identifiability_main} and Assumption \ref{Assumpt:NonDegeneracy} hold and assume that $ \sum_{t=1}^{T} \|P_t^{\ast} - \bar{P}^{\ast}\|_F^2 \le p^{\alpha} K$ for some $\alpha\in[0,1)$. For each $\delta\in(0,1)$, define the threshold
\begin{equation*}
    \tau_{n,p}
  = \max\limits_{1\le t\le T} \lrcp{\sqrt{n_t}}
    p^{\alpha/2} K^{1/2}
  + \sqrt{2\max\{1-l_p,u_p\} \max\{n,p\} \log\lrp{\frac{n + p}{\delta}}} 
  + \frac{2}{3} \log\lrp{\frac{n + p}{\delta}}.
\end{equation*}
If $\min\limits_{1\le t\le T} \lrcp{\underline{c}_{W_t^{\ast}} \sqrt{n_t}} \lrp{  \min\limits_{1\le t\le T} \lrcp{\underline{c}_{P_t^{\ast}} \sqrt{p}} - T^{-1/2} p^{\alpha/2} K^{1/2}} > 2 \tau_{n,p}$, then $\hat{K}=K$ holds with probability at least $1-\delta$.
\end{thm}

Unlike LDA and its variants \citep{blei2003latent, blei2006dynamic, mcauliffe2007supervised, srivastava2017autoencoding, bai2018neural, sridhar2022heterogeneous}, where theoretical guarantees for selecting $K$ are limited, our framework allows consistent rank recovery under explicit signal-to-noise conditions. Theorem \ref{Thm:RankConsistency} parallels rank-consistency results for static models \citep{klopp2023assigning, ke2024using}, but additional care is required here due to temporal drift and the decoupled occurrence–repetition structure. The main difficulty arises from the dynamic deviation term  $\sum_{t=1}^{T} \|P_t^{\ast}-\bar{P}^{\ast}\|_F^2$, which perturbs the common low-rank signal, and consistency holds only when this deviation is sufficiently small.

\section{Numerical Studies}\label{Sec:NumericalStudies}

We compare the proposed method with LDA-based \citep{blei2003latent, blei2006dynamic} and pLSI-based \citep{klopp2023assigning, ke2024using} baselines on synthetic and real dynamic corpora. 
Section \ref{Sec:NumericalStudies_Synthetic} evaluates finite-sample accuracy under controlled ground truth, and Section \ref{Sec:NumericalStudies_Real} reports empirical performance on the trimmed ICLR corpus with moving time windows.

\subsection{Simulated Data Analysis}\label{Sec:NumericalStudies_Synthetic}

We conduct synthetic experiments calibrated from the trimmed ICLR abstracts corpus \citep{gonzalez2024learning} to mimic realistic topic-word behavior. Each document in the dataset is an abstract associated with metadata and a manually assigned topic label.

We consider moving windows of length $T\in\{3,6\}$ ending in 2024 and restrict to the top $K\in\{3,4,5\}$ topics in the final year. The design aims to isolate the effects of key modeling components, including the separation of word occurrence and repetition, as well as temporal smoothness. Specifically, true topic-word occurrence and repetition parameters $(P_t^{\ast}, A_t^{\ast})$ are constructed under two regimes: (i) \emph{aligned}: high occurrence tends to imply high repetition; and (ii) \emph{misaligned}: repetition intensities deliberately permuted among frequent and infrequent words. Temporal drift is controlled by $\sigma\in\{0,0.3,0.6,0.9\}$, where larger values correspond to stronger temporal drift. Document-topic proportions follow Dirichlet designs with parameters $\{\Theta_0,\Theta_1\}$, corresponding to the cases when the active topic numbers $K_t$ remain unchanged over time or not. Documents are generated from the H-Multinomial model with $n_t=100$ per time point and repetition scaling $\rho\in\{0.25,1.0\}$; see Appendix \ref{Sec:NumericalStudies_Setup} for detailed experimental settings.

For comparison, we include DTM \citep{blei2006dynamic}, LDA \citep{blei2003latent} from the LDA family, and Topic-SCORE \citep{ke2024using}, SPOC \citep{klopp2023assigning} from the pLSI-family. Static methods are fitted on the pooled corpus and evaluated by year. We fix $K$ to its true value and report document-topic error $\mbox{Err}(\hat{W},W^{\ast})$ given by
\begin{equation*}
    \mbox{Err}(\hat{W}, W^{\ast})
  = \frac{1}{n} \sum\limits_{t=1}^{T} \sum\limits_{i=1}^{n_t}
    \lrp{\sum\limits_{k=1}^{K} (\hat{w}_{tik} - w_{tik}^{\ast})^2}^{1/2}.
\end{equation*}
All results are averaged over $M=20$ Monte Carlo replicates. LDA and DTM are additionally averaged over five random initializations to mitigate sensitivity to initialization.

\begin{figure}[!h]
    \centering
    \includegraphics[width=.95\linewidth]{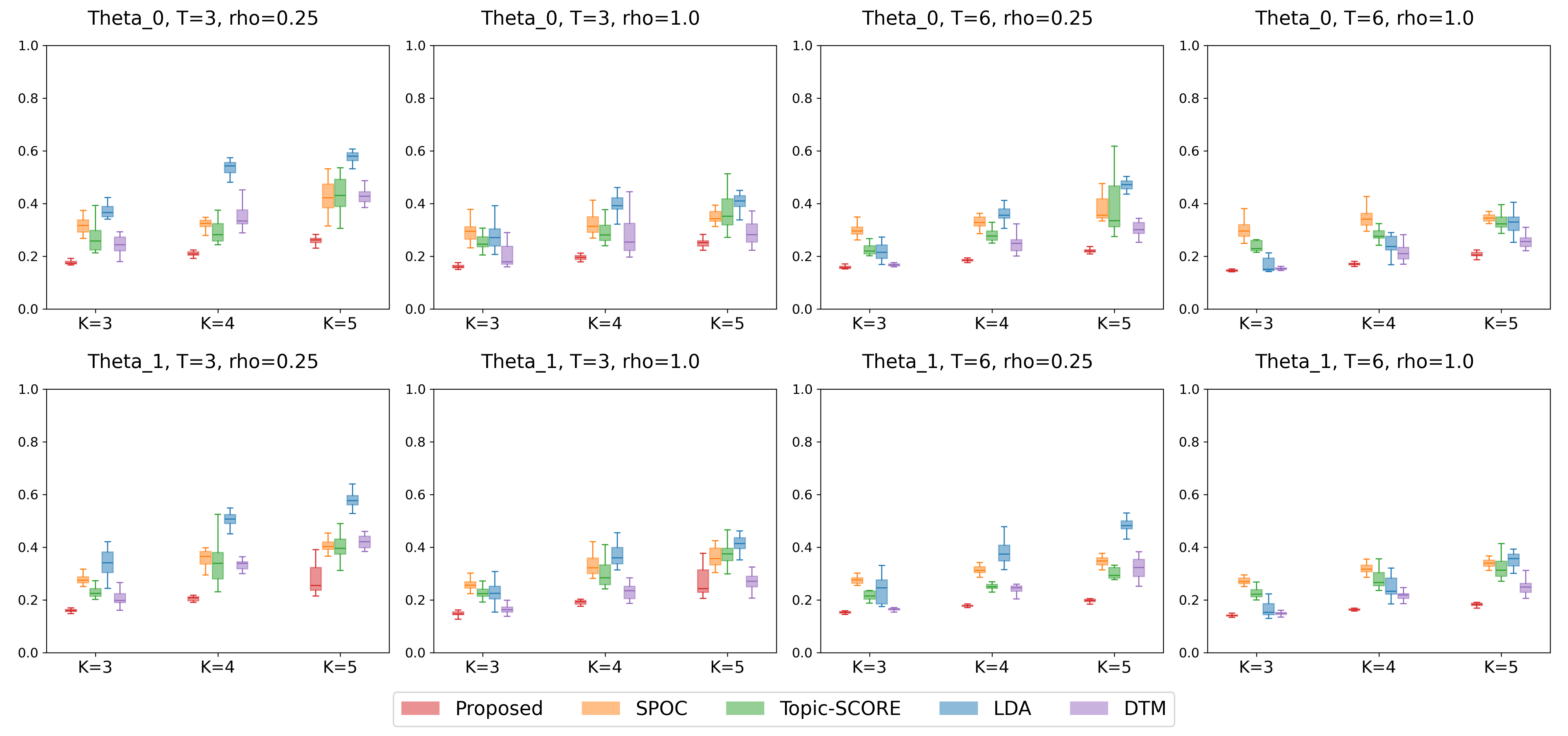}
    \caption{Boxplots of the document–topic estimation error $\mbox{Err}(\hat{W},W^{\ast})$ under the aligned setting with mixing parameter $\sigma=0.3$.}
    \label{Fig:Boxplot_W_error_v5_mix3}
\end{figure}

\begin{figure}[!h]
    \centering
    \includegraphics[width=.95\linewidth]{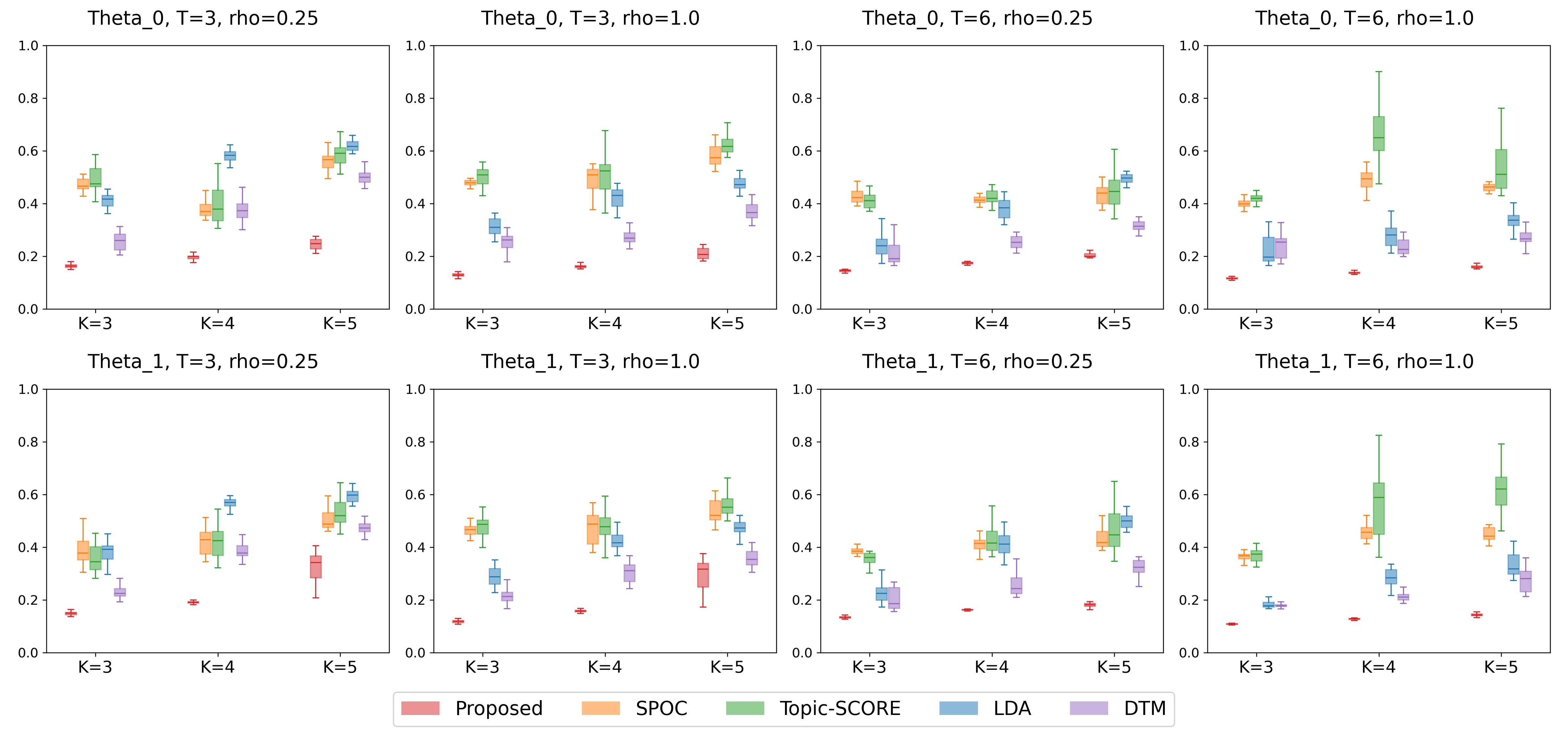}
    \caption{Boxplots of the document–topic estimation error $\mbox{Err}(\hat{W},W^{\ast})$ under the misaligned setting with mixing parameter $\sigma=0.3$.}
    \label{Fig:Boxplot_W_error_v4_mix3}
\end{figure}

The boxplots of $\mbox{Err}(\hat{W},W^{\ast})$ in Figures \ref{Fig:Boxplot_W_error_v5_mix3}-\ref{Fig:Boxplot_W_error_v4_mix3} show that the proposed method achieves the lowest estimation error under weak drift ($\sigma=0.3$), with the largest gains in the misaligned regime. 
When occurrence and repetition are aligned, multinomial-based competitors remain competitive as the hybrid likelihood approaches a multinomial model for sufficiently long documents. As expected, multinomial-based methods such as SPOC and Topic-SCORE perform competitively in this regime. 
When misaligned, collapsing the two signals into a single multinomial mechanism loses distinguishable information, whereas the proposed model preserves complementary structure and improves recovery by modeling word occurrence and repetition intensities separately. This observation provides direct empirical support for the decoupled modeling framework introduced in Section \ref{Sec:Representation}.

We note that variability increases moderately when $T$ is short, $K$ is large, and the active topic numbers vary ($\Theta_1$). As $K$ increases, overlap among topic–word profiles becomes more pronounced, whereas the distinguishable information is insufficient when each topic is observed only within limited windows. Notably, this variability is substantially reduced as $T$ increases and more observations per topic become available. We also observe that the performance of our method is relatively insensitive to $\rho$, while multinomial competitors benefit from larger effective document length, which is coherent with theoretical findings in \cite{klopp2023assigning} and \cite{ke2024using}.

\begin{figure}[!h]
    \centering
    \includegraphics[width=.95\linewidth]{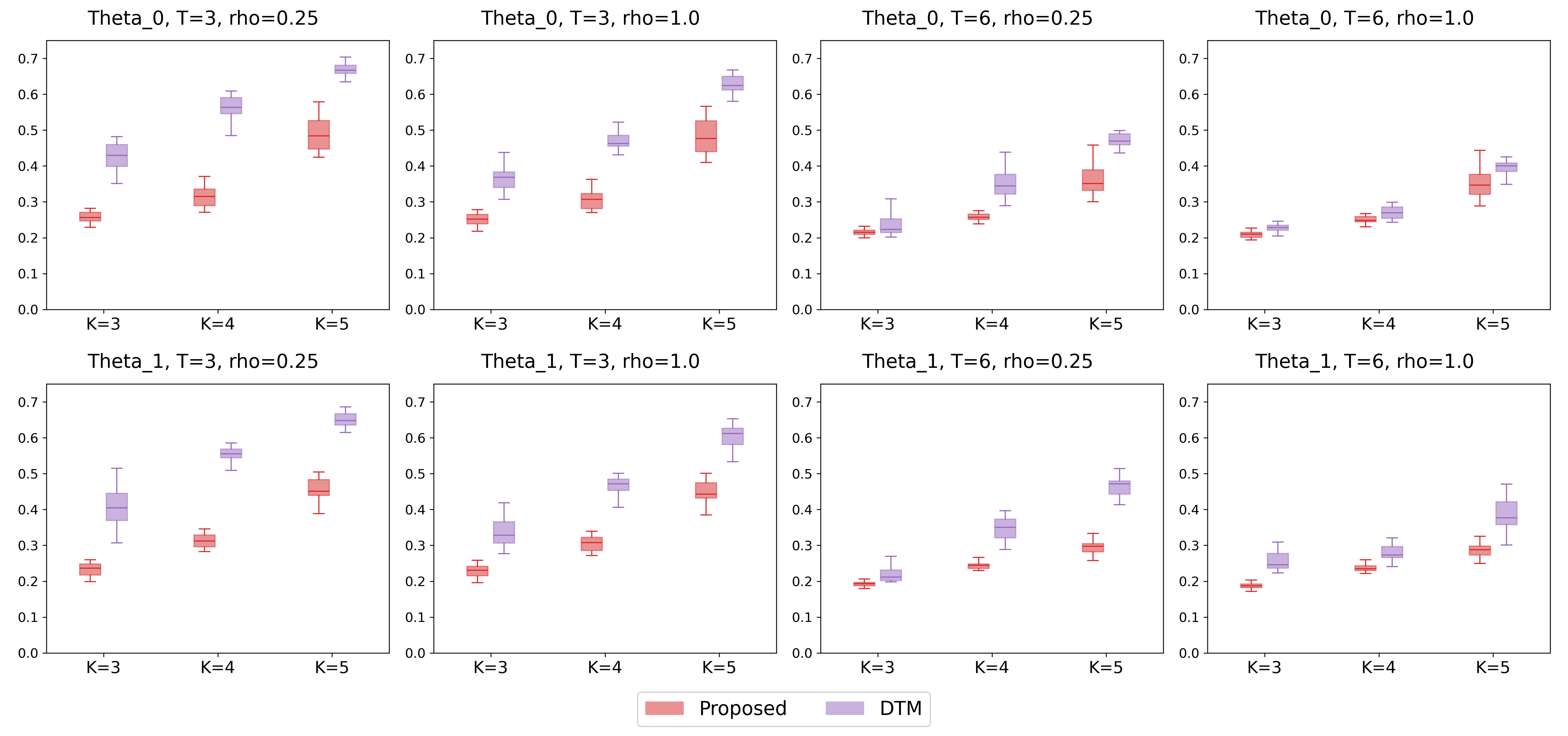}
    \caption{Boxplots of the document–topic estimation error $\mbox{Err}(\hat{W},W^{\ast})$ under the misaligned setting with strong temporal drift $\sigma=0.9$.}
    \label{Fig:Boxplot_W_error_v4_mix9}
\end{figure}

Under strong drift ($\sigma=0.9$), Figure \ref{Fig:Boxplot_W_error_v4_mix9} shows that the proposed method remains more accurate than DTM with comparable variability. In this regime, the temporal evolution of topic–word behavior is substantial. Unlike the state-space prior formulation with time-invariant prior mean used in \cite{blei2006dynamic}, our framework allows more flexible deviations by importing direct temporal regularization on both occurrence and repetition profiles. The advantage in this scenario demonstrates the robustness and flexibility of our model in challenging dynamic settings. 
Additional simulation results are provided in Appendix \ref{Sec:AddNumericalStudies_Synthetic}, covering the full grid of $(T, K, \sigma, \rho, \Theta)$ under both aligned and misaligned repetition settings, as well as estimation error analyses for the topic–word profiles $(P_t, A_t)$.

\subsection{Real Data Illustration}\label{Sec:NumericalStudies_Real}

We evaluate the proposed method on the trimmed ICLR corpus with moving windows of length $T\in\{3,4,5,6\}$ ending in different years. 
Within each window, we restrict to documents whose labels fall among the top $K\in\{3,4,5\}$ topics in the final year. The most prevalent topics from 2022 to 2024 are summarized in Table \ref{Tab:TopK_by_year}.

\begin{table}[H]
    \centering
    \caption{Most prevalent topic labels in the trimmed ICLR dataset from 2022 to 2024.}
    \label{Tab:TopK_by_year}
    \renewcommand{\arraystretch}{.65}
    \scalebox{0.85}{
    \begin{tabular}{c||c|c|c|c|c}
    \hline\hline
        Year & Top-1 & Top-2 & Top-3 & Top-4 & Top-5 \\ \hline
        2022 & RL & GANs & graphs & optimization & LLMs \\ \hline
        2023 & RL & graphs & LLMs & GANs & optimization \\ \hline
        2024 & LLMs & diffusion models & RL & graphs & optimization \\
    \hline\hline
    \end{tabular}}
\end{table}

We evaluate topic membership by the weighted $F_1$ score. Specifically, we assign the label to each document as its most dominant topic based on the estimates, that is, $\hat{g}_{it} = \argmax_{k\in\{1,\cdots,K\}} \hat{w}_{tik}$. Then we measure the agreement between $\hat{g}_{ti}$ and the true label. Additional basics of the trimmed dataset can be found in Appendix \ref{Sec:NumericalStudies_Setup_Basics}.

Two scenarios are considered: (i) we fix $T=3$ and vary the final year over $\{2022,2023,2024\}$; (ii) we fix the final year to 2024 and vary the window length $T\in\{3,4,5,6\}$. To isolate estimation performance, $K$ is fixed as the ground truth for all methods.

\begin{table}[!h]
    \centering
    \caption{Comparison of weighted $F_1$ score for the estimated topic membership on the trimmed ICLR corpus. Results are averaged over $M=5$ random initializations for LDA and DTM. The columns ``Imp(\%)'' reports the relative improvement (in percent) of the proposed method over each competing method.}
    \label{Tab:Real_Combined} 
    \renewcommand{\arraystretch}{.65}
    \scalebox{0.85}{
    \begin{tabular}{c|c||c|cc|cc|cc|cc}
    \hline\hline
        \multicolumn{11}{c}{\multirow{2}{*}{\textbf{Scenario I: $T=3$, varying last year}}} \\
        \multicolumn{11}{c}{} \\ \hline
        \multirow{2}{*}{Last year} & \multirow{2}{*}{$K$} & \multirow{2}{*}{Proposed} & \multicolumn{2}{c|}{SPOC} & \multicolumn{2}{c|}{Topic-SCORE} & \multicolumn{2}{c|}{LDA} & \multicolumn{2}{c}{DTM} \\ \cline{4-11}
        & & & $F_1$ & Imp(\%) & $F_1$ & Imp(\%) & $F_1$ & Imp(\%) & $F_1$ & Imp(\%) \\ \hline
        \multirow{3}{*}{2022} & 3 & 0.937 & 0.903 & 3.686 & 0.910 & 2.983 & 0.906 & 3.389 & 0.816 & 14.742 \\ \cline{2-11}
        & 4 & 0.873 & 0.616 & 41.869 & 0.830 & 5.190 & 0.814 & 7.228 & 0.798 & 9.406 \\ \cline{2-11}
        & 5 & 0.830 & 0.658 & 26.233 & 0.719 & 15.485 & 0.652 & 27.302 & 0.765 & 8.494 \\ \hline
        \multirow{3}{*}{2023} & 3 & 0.940 & 0.868 & 8.306 & 0.891 & 5.504 & 0.914 & 2.793 & 0.928 & 1.262 \\ \cline{2-11}
        & 4 & 0.882 & 0.806 & 9.484 & 0.815 & 8.259 & 0.724 & 21.873 & 0.780 & 13.173 \\ \cline{2-11}
        & 5 & 0.826 & 0.677 & 21.981 & 0.730 & 13.059 & 0.688 & 19.937 & 0.763 & 8.269 \\ \hline
        \multirow{3}{*}{2024} & 3 & 0.913 & 0.875 & 4.331 & 0.859 & 6.239 & 0.819 & 11.450 & 0.880 & 3.803 \\ \cline{2-11}
        & 4 & 0.901 & 0.859 & 4.928 & 0.849 & 6.126 & 0.805 & 12.012 & 0.884 & 1.896 \\  \cline{2-11}
        & 5 & 0.836 & 0.759 & 10.229 & 0.791 & 5.668 & 0.710 & 17.734 & 0.825 & 1.398 \\ \hline
        
        \multicolumn{11}{c}{\multirow{2}{*}{\textbf{Scenario II: last year 2024, varying $T$}}} \\
        \multicolumn{11}{c}{} \\ \hline
        \multirow{2}{*}{$T$} & \multirow{2}{*}{$K$} & \multirow{2}{*}{Proposed} & \multicolumn{2}{c|}{SPOC} & \multicolumn{2}{c|}{Topic-SCORE} & \multicolumn{2}{c|}{LDA} & \multicolumn{2}{c}{DTM} \\ \cline{4-11}
        & & & $F_1$ & Imp(\%) & $F_1$ & Imp(\%) & $F_1$ & Imp(\%) & $F_1$ & Imp(\%) \\ \hline
        \multirow{3}{*}{4} & 3 & 0.916 & 0.842 & 8.720 & 0.864 & 5.972 & 0.810 & 13.053 & 0.834 & 9.767 \\ \cline{2-11}
        & 4 & 0.897 & 0.811 & 10.592 & 0.850 & 5.587 & 0.841 & 6.772 & 0.765 & 17.314 \\ \cline{2-11}
        & 5 & 0.844 & 0.685 & 23.327 & 0.797 & 5.980 & 0.767 & 10.117 & 0.800 & 5.535 \\ \hline
        \multirow{3}{*}{5} & 3 & 0.899 & 0.812 & 10.734 & 0.858 & 4.733 & 0.838 & 7.286 & 0.729 & 23.274 \\\cline{2-11}
        & 4 & 0.897 & 0.715 & 25.440 & 0.842 & 6.551 & 0.754 & 18.850 & 0.851 & 5.421 \\ \cline{2-11}
        & 5 & 0.857 & 0.682 & 25.715 & 0.802 & 6.908 & 0.725 & 18.253 & 0.820 & 4.605 \\ \hline
        \multirow{3}{*}{6} & 3 & 0.910 & 0.689 & 32.015 & 0.846 & 7.545 & 0.752 & 20.961 & 0.846 & 7.589 \\ \cline{2-11}
        & 4 & 0.885 & 0.693 & 27.650 & 0.841 & 5.248 & 0.809 & 9.369 & 0.811 & 9.107 \\ \cline{2-11}
        & 5 & 0.853 & 0.640 & 33.374 & 0.791 & 7.850 & 0.666 & 28.165 & 0.745 & 14.477 \\
    \hline\hline
    \end{tabular}}
\end{table}

In Table \ref{Tab:Real_Combined}, we report both the raw weighted $F_1$ scores and the relative improvement rate (``Imp'') of the proposed method over each competitor. Across the table, the proposed method consistently achieves the highest weighted $F_1$ score. The relative improvements exceed 5\% in most cases and surpass 20\% in several settings with larger $K$. 
This aligns with the synthetic findings: when topics overlap in marginal frequencies, higher-order structure in occurrence-repetition patterns becomes increasingly informative for separating topics, and our method gains estimation accuracy by preserving the higher-order word interactions in both signals.
Moreover, the $F_1$ scores of our method remain stable across different moving windows, whereas DTM and LDA exhibit greater sensitivity to initialization.

As the temporal span increases, semantic evolution within topics accumulates, making accurate topic membership estimation more challenging. Our method is designed to accommodate such dynamic changes through direct temporal regularization on topic–word profiles. Consequently, both word occurrence and repetition intensities are allowed to evolve flexibly over time, subject to penalized deviation from the temporal average. This mechanism yields increasingly stable and accurate topic membership estimation as $T$ grows.

In contrast, static methods such as SPOC and LDA do not explicitly model temporal dynamics, resulting in gradually declining accuracy as $T$ increases. As discussed in Section \ref{Sec:NumericalStudies_Synthetic}, DTM captures temporal evolution through time-invariant prior parameters but remains sensitive to initialization and does not consistently benefit from longer windows. Topic-SCORE, which relies on geometric properties derived from representative words, exhibits relatively robust performance under moderate increases in $T$. This suggests that, over sufficiently long time spans, certain representative words retain distinguishable marginal proportions.
Nevertheless, the superior performance suggests that informative structure beyond marginal word frequencies that is captured through higher-order co-occurrence patterns in our framework remains essential for accurate topic recovery. The qualitative visualization of the recovered topics is presented in Appendix \ref{Sec:AddNumericalStudies_Real}.

\section{Conclusion}\label{Sec:Conclusion}

In this article, we proposed a hypergraph-based representation for text corpora that departs from the classical BOW–multinomial paradigm. The key innovation is the explicit separation of word occurrence and repetition intensities through document-level support and node-specific weights. This formulation induces a Bernoulli–Multinomial factorization with support-dependent normalization, allowing higher-order word interactions to be retained while preserving a likelihood-based framework amenable to statistical analysis.

Building on this representation, we developed a dynamic topic modeling framework based on structured low-rank factorizations of the occurrence and repetition components. The proposed estimator is computed via projected gradient descent and is supported by rigorous theoretical guarantees, including local linear convergence and non-asymptotic error bounds. These results extend likelihood-based topic modeling theory beyond the standard multinomial setting through novel arguments for perturbation analyses. Empirical studies demonstrate that modeling occurrence and repetition separately improves topic discrimination, particularly when semantic themes overlap. This supports the broader perspective that richer dependence structures can enhance identifiability and estimation in dynamic text analysis.

More broadly, the proposed hypergraphical likelihood framework provides a principled approach for modeling structured discrete data with heterogeneous support and intensity patterns, and may be useful beyond dynamic topic modeling.
For example, in mobile application logs, user activity data records both whether a user engages with a particular function (activation support) and how frequently or how long that function is used (intensity). Similarly, investment transaction records contain whether an investor participates in a given asset (activation support) and the amount or frequency of investment (intensity). In both settings, the support and intensity mechanisms are conceptually distinct and may exhibit different dependence structures across individuals or items. The hypergraphical representation, together with the explicit separation of activation and intensity components, provides a natural framework for modeling such heterogeneous behavioral patterns and is a promising direction for future research.


\section{Data Availability Statement}\label{data-availability-statement}
The raw ICLR dataset used in Section \ref{Sec:NumericalStudies} is publicly available at \url{https://github.com/berenslab/iclr-dataset}; see also \cite{gonzalez2024learning}. The detailed preprocessing procedures are described in the supplementary materials.

\phantomsection\label{supplementary-material}
\bigskip

\begin{center}

{\large\bf SUPPLEMENTARY MATERIAL}

\end{center}

\begin{description}

\item[Title:]

\textbf{Supplement to ``Dynamic Topic Modeling with a Higher-Order Hypergraphical Representation"} 

This supplement contains: (i) a heuristic initialization procedure, (ii) implementation and experimental details, (iii) additional numerical results, (iv) complete formulations of assumptions and technical conditions, and (v) full proofs of all theoretical results.

Appendix \ref{Sec:Method_Init} describes a heuristic initialization procedure for Algorithm \ref{Alg:PGD}. 
Appendices \ref{Sec:NumericalStudies_Setup}-\ref{Sec:AddNumericalStudies} provide detailed simulation settings and additional experimental results supplementing Section \ref{Sec:NumericalStudies} of the main article. 
Appendix \ref{Sec:FullNotationAssumptions} presents the complete formulations of identifiability, local regularity, and initialization conditions. 
Appendix \ref{Sec:ProofMain} contains the proofs of all main theorems, whereas Appendices \ref{Sec:Technical_Lemma}-\ref{Sec:Auxiliary_Lemma} present all the lemmas.

\end{description}

\bibliographystyle{agsm}
\bibliography{ref}

@article{ye2024dynamic,
  title={Dynamic topic language model on heterogeneous children’s mental health clinical notes},
  author={Ye, Hanwen and Moreno, Tatiana and Alpern, Adrianne and Ehwerhemuepha, Louis and Qu, Annie},
  journal={The Annals of Applied Statistics},
  volume={18},
  number={4},
  pages={3165--3184},
  year={2024},
  publisher={Institute of Mathematical Statistics}
}

@article{mcauliffe2007supervised,
  title={Supervised topic models},
  author={Mcauliffe, Jon and Blei, David},
  journal={Advances in neural information processing systems},
  volume={20},
  year={2007}
}

@article{blei2006correlated,
  title={Correlated topic models},
  author={Blei, David and Lafferty, John},
  journal={Advances in neural information processing systems},
  volume={18},
  pages={147},
  year={2006},
  publisher={MIT; 1998}
}

@article{sridhar2022heterogeneous,
  title={Heterogeneous supervised topic models},
  author={Sridhar, Dhanya and Daum{\'e} III, Hal and Blei, David},
  journal={Transactions of the Association for Computational Linguistics},
  volume={10},
  pages={732--745},
  year={2022},
  publisher={MIT Press One Broadway, 12th Floor, Cambridge, Massachusetts 02142, USA~…}
}

@article{roberts2014structural,
  title={Structural topic models for open-ended survey responses},
  author={Roberts, Margaret E and Stewart, Brandon M and Tingley, Dustin and Lucas, Christopher and Leder-Luis, Jetson and Gadarian, Shana Kushner and Albertson, Bethany and Rand, David G},
  journal={American journal of political science},
  volume={58},
  number={4},
  pages={1064--1082},
  year={2014},
  publisher={Wiley Online Library}
}

@article{sobhani2024multi,
  title={Multi-environment Topic Models},
  author={Sobhani, Dominic and Feder, Amir and Blei, David},
  journal={arXiv preprint arXiv:2410.24126},
  year={2024}
}

@article{chen2020modeling,
  title={Modeling multiplexed images with spatial-LDA reveals novel tissue microenvironments},
  author={Chen, Zhenghao and Soifer, Ilya and Hilton, Hugo and Keren, Leeat and Jojic, Vladimir},
  journal={Journal of Computational Biology},
  volume={27},
  number={8},
  pages={1204--1218},
  year={2020},
  publisher={Mary Ann Liebert, Inc., publishers 140 Huguenot Street, 3rd Floor New~…}
}

@article{blei2003latent,
  title={Latent dirichlet allocation},
  author={Blei, David M and Ng, Andrew Y and Jordan, Michael I},
  journal={Journal of machine Learning research},
  volume={3},
  number={Jan},
  pages={993--1022},
  year={2003}
}

@inproceedings{blei2006dynamic,
  title={Dynamic topic models},
  author={Blei, David M and Lafferty, John D},
  booktitle={Proceedings of the 23rd international conference on Machine learning},
  pages={113--120},
  year={2006}
}

@article{klopp2023assigning,
  title={Assigning topics to documents by successive projections},
  author={Klopp, Olga and Panov, Maxim and Sigalla, Suzanne and Tsybakov, Alexandre B},
  journal={The Annals of Statistics},
  volume={51},
  number={5},
  pages={1989--2014},
  year={2023},
  publisher={Institute of Mathematical Statistics}
}

@article{ke2024using,
  title={Using SVD for topic modeling},
  author={Ke, Zheng Tracy and Wang, Minzhe},
  journal={Journal of the American Statistical Association},
  volume={119},
  number={545},
  pages={434--449},
  year={2024},
  publisher={Taylor \& Francis}
}

@inproceedings{hofmann1999probabilistic,
  title={Probabilistic latent semantic analysis.},
  author={Hofmann, Thomas and others},
  booktitle={UAI},
  volume={99},
  pages={289--296},
  year={1999}
}

@article{jung2024graph,
  title={Graph-Structured Topic Modeling for Documents with Spatial or Covariate Dependencies},
  author={Jung, Yeo Jin and Donnat, Claire},
  journal={arXiv preprint arXiv:2412.14477},
  year={2024}
}

@inproceedings{arora2012learning,
  title={Learning topic models--going beyond SVD},
  author={Arora, Sanjeev and Ge, Rong and Moitra, Ankur},
  booktitle={2012 IEEE 53rd annual symposium on foundations of computer science},
  pages={1--10},
  year={2012},
  organization={IEEE}
}

@inproceedings{arora2013practical,
  title={A practical algorithm for topic modeling with provable guarantees},
  author={Arora, Sanjeev and Ge, Rong and Halpern, Yonatan and Mimno, David and Moitra, Ankur and Sontag, David and Wu, Yichen and Zhu, Michael},
  booktitle={International conference on machine learning},
  pages={280--288},
  year={2013},
  organization={PMLR}
}

@article{ke2019community,
  title={Community detection for hypergraph networks via regularized tensor power iteration},
  author={Ke, Zheng Tracy and Shi, Feng and Xia, Dong},
  journal={arXiv preprint arXiv:1909.06503},
  year={2019}
}

@article{wu2024general,
  title={A General Latent Embedding Approach for Modeling Non-uniform High-dimensional Sparse Hypergraphs with Multiplicity},
  author={Wu, Shihao and Xu, Gongjun and Zhu, Ji},
  journal={arXiv preprint arXiv:2410.12108},
  year={2024}
}

@article{gonzalez2024learning,
  title={Learning representations of learning representations},
  author={Gonz{\'a}lez-M{\'a}rquez, Rita and Kobak, Dmitry},
  journal={arXiv preprint arXiv:2404.08403},
  year={2024}
}

@article{holland1983stochastic,
  title={Stochastic blockmodels: First steps},
  author={Holland, Paul W and Laskey, Kathryn Blackmond and Leinhardt, Samuel},
  journal={Social networks},
  volume={5},
  number={2},
  pages={109--137},
  year={1983},
  publisher={Elsevier}
}

@article{hoff2002latent,
  title={Latent space approaches to social network analysis},
  author={Hoff, Peter D and Raftery, Adrian E and Handcock, Mark S},
  journal={Journal of the american Statistical association},
  volume={97},
  number={460},
  pages={1090--1098},
  year={2002},
  publisher={Taylor \& Francis}
}

@inproceedings{sarioglu2012clinical,
  title={Clinical report classification using natural language processing and topic modeling},
  author={Sarioglu, Efsun and Choi, Hyeong-Ah and Yadav, Kabir},
  booktitle={2012 11th international conference on machine learning and applications},
  volume={2},
  pages={204--209},
  year={2012},
  organization={IEEE}
}

@article{curiskis2020evaluation,
  title={An evaluation of document clustering and topic modelling in two online social networks: Twitter and Reddit},
  author={Curiskis, Stephan A and Drake, Barry and Osborn, Thomas R and Kennedy, Paul J},
  journal={Information Processing \& Management},
  volume={57},
  number={2},
  pages={102034},
  year={2020},
  publisher={Elsevier}
}

@article{palese2018relative,
  title={The relative importance of service quality dimensions in E-commerce experiences},
  author={Palese, Biagio and Usai, Antonio},
  journal={International Journal of Information Management},
  volume={40},
  pages={132--140},
  year={2018},
  publisher={Elsevier}
}

@article{yuan2018topic,
  title={Topic sentiment mining for sales performance prediction in e-commerce},
  author={Yuan, Hui and Xu, Wei and Li, Qian and Lau, Raymond},
  journal={Annals of Operations Research},
  volume={270},
  pages={553--576},
  year={2018},
  publisher={Springer}
}

@article{harris1954distributional,
  title={Distributional structure},
  author={Harris, Zellig S},
  journal={Word},
  volume={10},
  number={2-3},
  pages={146--162},
  year={1954},
  publisher={Taylor \& Francis}
}

@inproceedings{chong2009simultaneous,
  title={Simultaneous image classification and annotation},
  author={Chong, Wang and Blei, David and Li, Fei-Fei},
  booktitle={2009 IEEE Conference on computer vision and pattern recognition},
  pages={1903--1910},
  year={2009},
  organization={IEEE}
}

@article{fan2025covariates,
  title={Covariates-Adjusted Mixed-Membership Estimation: A Novel Network Model with Optimal Guarantees},
  author={Fan, Jianqing and Ge, Jiawei and Hou, Jikai},
  journal={arXiv preprint arXiv:2502.06671},
  year={2025}
}

@article{ma2020universal,
  title={Universal latent space model fitting for large networks with edge covariates},
  author={Ma, Zhuang and Ma, Zongming and Yuan, Hongsong},
  journal={Journal of Machine Learning Research},
  volume={21},
  number={4},
  pages={1--67},
  year={2020}
}

@inproceedings{yao2019graph,
  title={Graph convolutional networks for text classification},
  author={Yao, Liang and Mao, Chengsheng and Luo, Yuan},
  booktitle={Proceedings of the AAAI conference on artificial intelligence},
  volume={33},
  number={01},
  pages={7370--7377},
  year={2019}
}

@inproceedings{rousseau2015text,
  title={Text categorization as a graph classification problem},
  author={Rousseau, Fran{\c{c}}ois and Kiagias, Emmanouil and Vazirgiannis, Michalis},
  booktitle={Proceedings of the 53rd Annual Meeting of the Association for Computational Linguistics and the 7th International Joint Conference on Natural Language Processing (Volume 1: Long Papers)},
  pages={1702--1712},
  year={2015}
}

@article{ding2020more,
  title={Be more with less: Hypergraph attention networks for inductive text classification},
  author={Ding, Kaize and Wang, Jianling and Li, Jundong and Li, Dingcheng and Liu, Huan},
  journal={arXiv preprint arXiv:2011.00387},
  year={2020}
}

@inproceedings{rousseau2013graph,
  title={Graph-of-word and TW-IDF: new approach to ad hoc IR},
  author={Rousseau, Fran{\c{c}}ois and Vazirgiannis, Michalis},
  booktitle={Proceedings of the 22nd ACM international conference on Information \& Knowledge Management},
  pages={59--68},
  year={2013}
}

@article{pradeepa2024hgatt_lr,
  title={HGATT\_LR: transforming review text classification with hypergraphs attention layer and logistic regression},
  author={Pradeepa, S and Jomy, Elizabeth and Vimal, S and Hassan, Md Mehedi and Dhiman, Gaurav and Karim, Asif and Kang, Dongwann},
  journal={Scientific Reports},
  volume={14},
  number={1},
  pages={19614},
  year={2024},
  publisher={Nature Publishing Group UK London}
}

@article{bazaga2024hyperbert,
  title={HyperBERT: Mixing hypergraph-aware layers with language models for node classification on text-attributed hypergraphs},
  author={Bazaga, Adri{\'a}n and Li{\`o}, Pietro and Micklem, Gos},
  journal={arXiv preprint arXiv:2402.07309},
  year={2024}
}

@article{lyu2023optimal,
  title={Optimal clustering of discrete mixtures: binomial, Poisson, block models, and multi-layer networks},
  author={Lyu, Zhongyuan and Li, Ting and Xia, Dong},
  journal={arXiv preprint arXiv:2311.15598},
  year={2023}
}

@article{zhang2023generalized,
  title={Generalized connectivity matrix response regression with applications in brain connectivity studies},
  author={Zhang, Jingfei and Sun, Will Wei and Li, Lexin},
  journal={Journal of Computational and Graphical Statistics},
  volume={32},
  number={1},
  pages={252--262},
  year={2023},
  publisher={Taylor \& Francis}
}

@inproceedings{ma2018implicit,
  title={Implicit regularization in nonconvex statistical estimation: Gradient descent converges linearly for phase retrieval and matrix completion},
  author={Ma, Cong and Wang, Kaizheng and Chi, Yuejie and Chen, Yuxin},
  booktitle={International conference on machine learning},
  pages={3345--3354},
  year={2018},
  organization={PMLR}
}

@article{chi2019nonconvex,
  title={Nonconvex optimization meets low-rank matrix factorization: An overview},
  author={Chi, Yuejie and Lu, Yue M and Chen, Yuxin},
  journal={IEEE Transactions on Signal Processing},
  volume={67},
  number={20},
  pages={5239--5269},
  year={2019},
  publisher={IEEE}
}

@article{chen2015fast,
  title={Fast low-rank estimation by projected gradient descent: General statistical and algorithmic guarantees},
  author={Chen, Yudong and Wainwright, Martin},
  journal={arXiv preprint arXiv:1509.03025},
  year={2015}
}

@article{chen2018projected,
  title={The projected power method: An efficient algorithm for joint alignment from pairwise differences},
  author={Chen, Yuxin and Cand{\`e}s, Emmanuel J},
  journal={Communications on Pure and Applied Mathematics},
  volume={71},
  number={8},
  pages={1648--1714},
  year={2018},
  publisher={Wiley Online Library}
}

@inproceedings{bai2018neural,
  title={Neural relational topic models for scientific article analysis},
  author={Bai, Haoli and Chen, Zhuangbin and Lyu, Michael R and King, Irwin and Xu, Zenglin},
  booktitle={Proceedings of the 27th ACM International Conference on Information and Knowledge Management},
  pages={27--36},
  year={2018}
}

@article{srivastava2017autoencoding,
  title={Autoencoding variational inference for topic models},
  author={Srivastava, Akash and Sutton, Charles},
  journal={arXiv preprint arXiv:1703.01488},
  year={2017}
}

\end{document}